\documentclass[12pt]{article}

\usepackage{preprint}

\usepackage{amsmath, amsthm, amssymb, amsfonts}

\usepackage[numbers,square]{natbib}
\usepackage[utf8]{inputenc}
\usepackage[T1]{fontenc}
\usepackage{xcolor}
\usepackage[colorlinks=true,
            linkcolor=purple,
            urlcolor=blue,
            citecolor=cyan,
            anchorcolor=black]{hyperref}
\usepackage{booktabs}
\usepackage{nicefrac}
\usepackage{microtype}
\usepackage{float}
\usepackage{graphicx}
\usepackage{multirow}
\usepackage{array}
\providecommand{\arraybackslash}{\let\\=\tabularnewline}

\newcommand{\ecfrsection}{https://www.ecfr.gov/current/title-45/section-164.514}
\newcommand{\cfr}[1]{\href{\ecfrsection}{#1}}

\usepackage{setspace}
\usepackage{newfloat}
\DeclareFloatingEnvironment[name={Supplementary Figure}]{suppfigure}
\usepackage{sidecap}
\sidecaptionvpos{figure}{c}

\usepackage{titlesec}
\titleformat{\section}[block]{\large\bfseries\raggedright}{}{0em}{\MakeUppercase}
\titleformat{\subsection}[block]{\normalsize\bfseries\raggedright}{}{0em}{}
\titleformat{\subsubsection}[block]{\normalsize\mdseries\raggedright}{}{0em}{}
\titleformat{\paragraph}[runin]{\normalsize\itshape}{}{0em}{}
\titlespacing\section{0pt}{14pt plus 4pt minus 3pt}{4pt plus 2pt minus 1pt}
\titlespacing\subsection{0pt}{12pt plus 3pt minus 3pt}{2pt plus 1pt minus 1pt}
\titlespacing\subsubsection{0pt}{10pt plus 3pt minus 3pt}{1pt plus 1pt minus 1pt}

\usepackage{tikz}

\newcommand{\phitag}[1]{\texttt{[#1]}}

\newcommand{\prm}[1]{\emph{#1}}

\title{Institution-Specific LLM Prompting Recovers PHI That De-identification
Systems and Their Gold Standards Both Miss}
\date{}

\SetBgContents{}

\begin{document}

\thispagestyle{empty}

\begin{center}
{\LARGE\bfseries Institution-Specific LLM Prompting Recovers PHI That
De-identification Systems and Their Gold Standards Both Miss\par}
\vspace{0.4cm}
{\small Article type: Research and Applications}
\end{center}

\vspace{0.4cm}

\noindent\textbf{Authors.}
Daniel Palacios, BS\textsuperscript{1,2,3,4,$\dagger$},
Matthew Brady Neeley, BS\textsuperscript{1,2,3,4,$\dagger$},
Angel Adetomike Otto, MS\textsuperscript{5},
Shalini Dhamodharan, MS\textsuperscript{5},
John P. Woodhouse, BA\textsuperscript{5},
Chi-fan Lin, MS\textsuperscript{5},
Mark Zobeck, MD, MPH\textsuperscript{5,$*$},
Zhandong Liu, PhD\textsuperscript{1,2,3,4,$*$},
Hyun-Hwan Jeong, PhD\textsuperscript{2,3,4,$*$}.

\smallskip
\noindent\textsuperscript{$\dagger$}These authors contributed equally.
\quad\textsuperscript{$*$}Co-corresponding authors.

\smallskip

\medskip
\noindent\textbf{Affiliations.}
\begin{description}\setlength{\itemsep}{0pt}
\item[1.] Quantitative and Computational Biosciences, Baylor College of Medicine, Houston, Texas, USA
\item[2.] Department of Pediatrics, Baylor College of Medicine, Houston, Texas, USA
\item[3.] Jan and Dan Duncan Neurological Research Institute, Texas Children's Hospital, Houston, Texas 77030, USA
\item[4.] Data Science Center, Texas Children's Hospital, Houston, Texas 77030, USA
\item[5.] Section of Hematology-Oncology, Department of Pediatrics, Baylor College of Medicine, Houston, Texas, USA
\end{description}

\medskip
\noindent\textbf{Corresponding author.}
Hyun-Hwan Jeong, Department of Pediatrics, Baylor College of Medicine, and Jan
and Dan Duncan Neurological Research Institute, Texas Children's Hospital, 1250
Moursund Street, Houston, TX 77030, USA. Telephone: +1~832-824-1000, ext.~25535.
Email: \texttt{hyun-hwan.jeong@bcm.edu}.

\medskip
\noindent\textbf{Keywords:}
data anonymization; electronic health records; natural language processing;
machine learning; large language models.

\medskip
\noindent\textbf{Word count.} Abstract: 239. Main body: 3{,}808. Tables: 1. Figures: 4.

\clearpage

\begin{center}\large\bfseries ABSTRACT\end{center}

\noindent\textbf{Objective.}
Secondary use of electronic health records requires de-identification, yet
existing systems miss \emph{institutionally situated} protected health information
(PHI): identifiers such as hospital abbreviations and building names whose status is
locally determined. We evaluated whether large language models (LLMs) can close this
gap through in-context learning while controlling precision and recall.

\smallskip
\noindent\textbf{Materials and Methods.}
On 100 annotated pediatric oncology notes (5,322 PHI spans) we benchmarked eight
LLMs against two purpose-built systems (Stanford TiDE, OpenMed PII) and two
pattern-based baselines. Each LLM was run in three prompt modes: \prm{Baseline}
(HIPAA-aligned), \prm{Targeted} (plus institutional PHI categories), and
\prm{Precision} (plus instructions against over-redaction). We also compared
14~multi-agent and ensemble configurations. Recall was the primary safety
metric.

\smallskip
\noindent\textbf{Results.}
LLMs outperformed the purpose-built systems (best F1=0.918$\pm$0.001, Sonnet~4.6,
vs.\ TiDE 0.779), with advantages concentrated in contextual categories. Naming the
missed categories recovered 79\% (48/61), and \prm{Precision} recovered precision.
No agentic architecture beat single-pass prompting (F1 0.906--0.908). The LLMs also
redacted 414~identifiers absent from the gold standard, scored false positive;
expert review of 49 confirmed all as true PHI, and re-annotating the
10~highest-discrepancy notes (+227~spans) lifted \prm{Precision} to recall=0.981
(F1=0.907$\pm$0.002).

\smallskip
\noindent\textbf{Discussion.}
Naming an institution's own identifiers and warning against over-redaction resolves
both the institutional PHI gap and the precision--recall trade-off in one LLM call
per note. LLMs can cost more, but that buys a way to audit the standard.

\smallskip
\noindent\textbf{Conclusion.}
LLMs are an adaptable alternative to purpose-built de-identification;
institution-specific prompt development should be the primary adaptation.

\clearpage

\section*{Introduction}
\label{sec:background}

Secondary use of electronic health records (EHRs) requires de-identification of
protected health information (PHI) under HIPAA's Safe Harbor standard, which
specifies 18 identifier categories~\cite{hipaa1996,ocr2012guidance}. These define
\emph{canonical} PHI (names, dates, medical record numbers, geographic data) that
models recognize from general linguistic patterns. Real clinical notes also
contain \textbf{institutionally situated PHI}: identifiers whose status depends on
local institutional context, including hospital abbreviations (``TCH'' for Texas
Children's Hospital), building names (``Mark Wallace Tower''), internal clinic
codes, and provider naming conventions unique to an institution. Although not enumerated among the 18 categories, they fall within HIPAA's
definition of PHI under both the Safe Harbor catch-all for ``any other unique
identifying number, characteristic, or code''
(\cfr{45~CFR~\S164.514(b)(2)(i)(R)}) and the expert-determination standard
(\cfr{\S164.514(b)(1)})~\cite{ocr2012guidance,cfr164514}: in a pediatric-oncology
population a named specialty facility, a rare diagnosis, and a service date can
jointly re-identify a record even after every canonical identifier is removed.
Each element is independently re-identifying: the set of facilities a patient
visits is itself a signature~\cite{malin2004trail,sweeney2000simple} and diagnosis
codes alone can breach privacy~\cite{loukides2010diagnosis}; in Washington State
discharge data carrying hospital, diagnosis, and attending physician but no names
or addresses, news reports uniquely matched 35 of 81 named patients to their
records~\cite{sweeney2013washington}. Removal is a regulatory requirement, not an
optional refinement.

A model prompted only on HIPAA's 18 categories has no basis to recognize that
``TCH'' is an identifying abbreviation or that a four-digit pager number is a staff
identifier. These are failures not of capability but of \emph{specification}: the
model was never told they require redaction. The OCR guidance anticipates this,
warning that esoteric notation such as acronyms known to only a few of a covered
entity's employees can lead to either unnecessary redaction or failure to
redact~\cite{ocr2012guidance}. Nor can the specification be written once and
reused, since note templates, abbreviations, and patient populations differ by
site~\cite{norgeot2020philter}: Veterans Health Administration notes required
customizing to institution-specific formats~\cite{ferrandez2012evaluating,meystre2014impact},
and cross-institute evaluations report consistent degradation on
transfer~\cite{yang2019crossinstitute}. The gap is acute in pediatric settings,
where large multidisciplinary teams author notes that reference caregivers and
carry institutional shorthand.

Recent work has begun isolating institution-level identifiers as a distinct
annotation class, notably the \textsc{hospital} category of SHIELD --- a recent
teacher--student distillation framework for clinical de-identification --- on which
both its teacher and student models record their lowest precision~\cite{shield2026}. No prior study
has characterized \emph{why} these identifiers fail, nor shown the failure
remediable through specification rather than retraining. Automated
de-identification has progressed from rule-based
systems~\cite{sweeney1996replacing,neamatullah2008automated} through neural
sequence models~\cite{dernoncourt2017identification,liu2017identification} to
transformer-based
NER~\cite{lee2020biobert,alsentzer2019publicly,paul2026deidclinicriskawarepseudonymizationframework},
benchmarked on i2b2/UTHealth shared
tasks~\cite{stubbs2015annotating,uzuner2007evaluating}. Two purpose-built systems
anchor our comparison: OpenMed's domain-adapted
NER~\cite{panahi2025openmedneropensourcedomainadapted} and Stanford's TiDE, which
combines NER, pattern matching, and known-PHI
lookup~\cite{datta2020newparadigmacceleratingclinical,callahan2023stanford} with
Hiding-in-Plain-Sight surrogates~\cite{carrell2013hiding}.

Large language models (LLMs) are competitive with or superior to traditional NER
systems on adult clinical benchmarks~\cite{liu2023deid_gpt}. Closest to this work,
Wiest et al.~\cite{wiest2025llmanonymizer} benchmarked eight local LLMs on 250
clinical letters, reporting $\sim$99.2\% PHI removal in a different language and
note type; Altalla' et al.~\cite{altalla2025evaluating} evaluated GPT-3.5 and
GPT-4 (P$\approx$0.99, R$\approx$0.83), and Pissarra et
al.~\cite{pissarra2024unlocking} found LLMs and Presidio baselines complementary. Multi-agent architectures have also been explored:
TEAM-PHI~\cite{team_phi_2025} ranks de-identification models with majority-voting
\emph{evaluation} agents and no gold labels, OEMA~\cite{oema2026} uses three
agents for zero-shot clinical NER, and
SHIELD~\cite{shield2026,gunay2024llmsinloop,kim2024generalizing} selects a teacher
labeler for distillation into locally-deployable
students~\cite{sounack2025bioclinicalmodernbertstateoftheartlongcontext}. The
prover-verifier framework~\cite{openai2024proververifier} justifies
generation-then-verification architectures, untested in de-identification.

A second challenge is the precision--recall
trade-off~\cite{buckland1994relationship}. De-identification has long favored
recall, since a missed identifier is a privacy breach whereas over-redaction only
removes clinical content~\cite{stubbs2015annotating,ferrandez2013bob}. Distinctive
in the single-pass LLM setting are its \emph{magnitude} and
\emph{model-dependence}: no single prompt optimizes both objectives across models,
and some over-redact severely enough to degrade data utility, an effect standard
metrics do not capture~\cite{aghakasiri2025doctorordered}. Conversely,
adversarial LLM-based re-identification~\cite{diri2024} shows even strong systems
leave notes vulnerable.

\label{sec:objective}

We hypothesized that what limits de-identification is \emph{specification}
rather than model capability: that naming an institution's own identifiers in
context would resolve both the institutional PHI gap and the precision--recall
trade-off. We tested this against the competing explanation that the trade-off
demands architectural remedy, using dual-pass and Scrubber--Auditor pipelines
drawn from the prover-verifier paradigm~\cite{openai2024proververifier} and
multi-agent clinical NLP~\cite{team_phi_2025} (Supplementary Note~3). We make three contributions.
First, benchmarking 8~LLMs against two purpose-built systems (Stanford TiDE,
OpenMed PII), pattern-based baselines, and multi-stage pipelines on
100~expert-annotated pediatric oncology notes (5,322~PHI spans), we introduce
\textbf{institutionally situated PHI} as a failure mode common to all. Second,
in-context learning adapts de-identification without fine-tuning: naming the missed
categories recovers most of them and anti-over-redaction instructions restore
precision, while error analysis exposed gold-standard gaps that expert re-annotation
confirmed as true PHI. Third, the trade-off resolves in a single pass: no agentic
architecture outperformed it on F1, locating the bottleneck in specification rather
than inference-time computation.

\section*{Methods}
\label{sec:methods}

\subsection*{Study design and clinical corpus}
\label{sec:study_design}

We benchmarked the LLMs and purpose-built systems listed in
Table~\ref{tab:models}, plus two pattern-based baselines (regex-only, and spaCy
NER + regex), on 100 pediatric oncology clinical notes, with the LLMs run under
each of the three prompt conditions below. No model was trained or fine-tuned.
Reporting follows TRIPOD-LLM~\cite{gallifant2025tripodllm};
Tables~S1 and~S2 give the
item-by-item adherence table.

\begin{figure}[htbp]
\centering
\includegraphics[width=\textwidth]{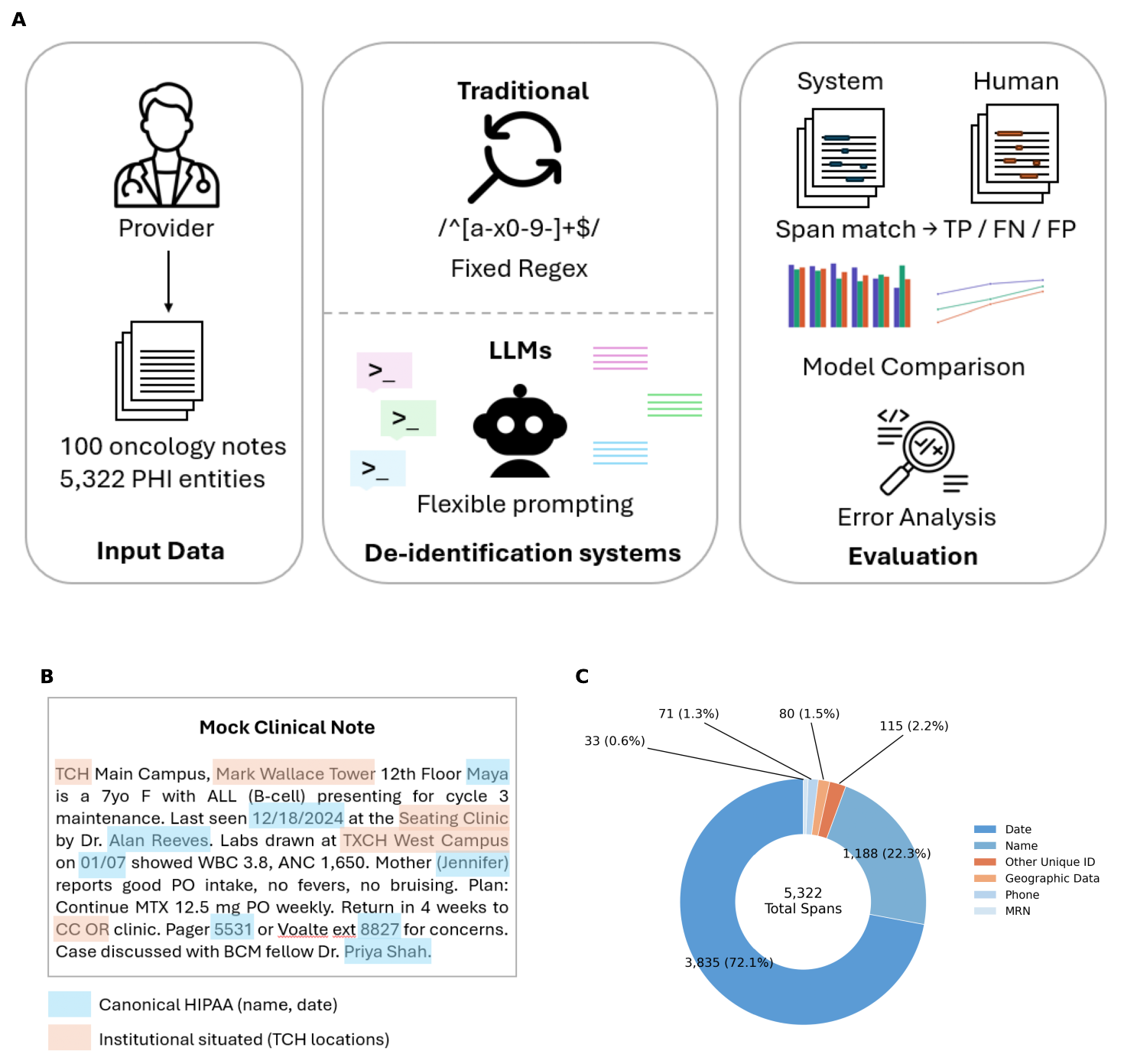}
\caption{Study design and corpus characteristics.
\textbf{(A)}~Study design: 8~LLMs (top~4 shown in main figures) +
traditional baselines evaluated on 100 pediatric oncology notes
(5,322 spans) under 3 prompt conditions.
\textbf{(B)}~Synthetic clinical note illustrating canonical HIPAA PHI
(names, dates; blue) vs.\ institutionally situated PHI (facility
abbreviations, building names; orange).
\textbf{(C)}~Gold standard PHI distribution over the six categories with
$N\geq20$ (four further categories hold $\leq5$ spans each; all ten, with their
denominators, are in Supplementary Table~S4): Date
(72.1\%) and Name (22.3\%) dominate, while institutionally enriched categories
(Other Unique ID, Geographic Data) comprise 3.7\% of spans. This distribution is
that of the \emph{original} gold standard; because that annotation under-counted
institutional identifiers (see Results), 3.7\% is a lower bound on their true
prevalence.}
\label{fig:figure1}
\end{figure}

\textbf{Clinical note corpus.} Our corpus consisted of 100 English-language
clinical notes from the pediatric oncology service at Texas Children's Hospital
(Houston, TX, USA; 96 patients), drawn by pseudorandom sort under a fixed seed from
notes dated on or after January 1, 2021, unstratified. Notes ranged from 91 to
32,767 characters (median, 6,504; mean, 10,058), five truncated at the export
limit. Three clinical annotators, trained by a pediatric oncologist and
informatician to identify both the 18 Safe Harbor categories and institutionally
situated identifiers, annotated 5,322 PHI spans across 97~notes (3 contained no
PHI), resolving questions with the oncologist and regulatory personnel; because
annotation followed consensus adjudication rather than independent double
annotation, no agreement statistic was computed. Figure~\ref{fig:figure1}C shows the category distribution
(Supplementary Note~1.1).

\textbf{Validation dataset.} To check that our findings were not corpus-specific,
we also evaluated all systems on 49~notes (409~gold PHI spans) from the USDHUB
repository, a separately curated pediatric neurology sample from the same
institution, independently de-identified by a different group, under the identical
\prm{Baseline} prompt and scoring pipeline (Supplementary Note~4). Because USDHUB
was built for TiDE and ships patient-specific provisioning material, TiDE was run
in two bracketing configurations: fully unprovisioned (identical footing to the
LLMs) and maximally provisioned with the supplied known-PHI dictionary and per-note
identifier header. For the provisioned run only, precision and F1 are scored on the
note body, since the injected header carries no gold spans (Supplementary
Note~4.1).

\subsection*{De-identification systems}
\label{sec:models}

\textbf{LLM models.} We evaluated eight LLMs via AWS Bedrock's Converse API
(Table~\ref{tab:models}) at temperature $= 0.0$ (except Opus~4.8, which does not
accept the parameter via Bedrock; Supplementary Note~3.3), max output tokens
$= 65{,}000$, and up to 3 retries with exponential backoff. For visual clarity,
main-text figures present the top~4 LLMs by \prm{Baseline} F1 (Sonnet~4.6,
Opus~4.8, GLM-5, DeepSeek~V3.2); all eight appear in
Tables~S3--S6.

\begin{table}[H]
\centering
\caption{LLMs and purpose-built systems evaluated. The two pattern-based
baselines (regex-only; spaCy NER + regex) are specified in the text and are
included in all reported comparisons.}
\label{tab:models}
\begin{tabular}{llll}
\toprule
\textbf{System} & \textbf{Type} & \textbf{Family} & \textbf{Parameters} \\
\midrule
Claude Opus 4.8   & LLM & Anthropic & Undisclosed \\
Claude Sonnet 4.6 & LLM & Anthropic & Undisclosed \\
GPT-oss-120B      & LLM & OpenAI    & 120B \\
GPT-oss-20B       & LLM & OpenAI    & 20B  \\
GLM-5             & LLM & Zhipu AI  & 754B \\
Kimi K2.5         & LLM & Moonshot  & 1.1T \\
MiniMax M2.5      & LLM & MiniMax   & 229B \\
DeepSeek V3.2     & LLM & DeepSeek  & 685B \\
\midrule
Stanford TiDE     & NER + Rules & Stanford NLP & --- \\
OpenMed PII       & NER (token classif.) & OpenMed & 434M \\
\bottomrule
\end{tabular}
\end{table}

\textbf{Stanford TiDE.} TiDE~\cite{datta2020newparadigmacceleratingclinical} is a
production system detecting PHI through NER, regex, and known-PHI matching against
a supplied list of each patient's real identifiers. Since the LLMs and OpenMed
never receive that list, input parity required running TiDE on note text alone,
with known-PHI matching disabled so field lookups find no rows while NER and regex
fire normally. This understates TiDE's production performance but isolates
contextual reasoning from pattern matching (Supplementary Note~4.1).

\textbf{OpenMed PII.} We used
OpenMed-PII-SuperClinical-Large-434M-v1~\cite{panahi2025openmedneropensourcedomainadapted,openmed_pii_modelcard},
a transformer token-classification model fine-tuned for personally identifiable
information (434M parameters, 54 sensitive-information types), whose bracketed
placeholders match the LLM output format, so the same scoring pipeline applies.

\subsection*{Prompt conditions}
\label{sec:prompts}

The \prm{Baseline} prompt asks the model to return the exact input text with all
18 HIPAA Safe Harbor categories replaced by typed placeholders, enumerating the
categories with examples. \prm{Targeted} appends four categories of institutionally
situated PHI that \prm{Baseline} error analysis surfaced: staff names adjacent to
credentials, pager and Voalte numbers, institution names and abbreviations
(``TCH'', ``TXCH''), and building or facility names. Department and clinic codes, a
fifth subcategory the same analysis surfaced, were \emph{not} named
(Table~S7). \prm{Precision} retains all of \prm{Targeted} and adds
a ``DO NOT OVER-REDACT'' block covering the six largest observed false-positive
categories, each with WRONG~$\rightarrow$~RIGHT pairs, closing with an instruction
to redact anyway when genuinely uncertain (Supplementary Note~2.2).

\subsection*{Multi-stage architectures}
\label{sec:multi_agent}

We tested whether architectural complexity could outperform single-pass prompt
engineering, using \emph{multi-stage} for any pipeline with more than one LLM call,
\emph{multi-agent} for pipelines whose calls occupy distinct roles (scrubber,
auditor, verifier), and \emph{heterogeneous} for the subset combining two or more
models. Three paradigms were explored: \textbf{dual-pass iterative refinement}
(Sonnet~4.6 applied twice, the second pass searching its own output for residual
institutional PHI), \textbf{heterogeneous Scrubber--Auditor} (a recall-maximizing
Sonnet~4.6 scrubber followed by a precision-focused Opus~4.8 auditor), and
\textbf{cross-model dual-pass} (two models with complementary error profiles).
Descriptions, prompts, and design rationale are in Supplementary Note~3. The
top~3 configurations were run for 5~independent trials with statistical
comparison (McNemar's exact test, Wilcoxon signed-rank, bootstrap 95\% CIs,
Holm--Bonferroni corrected across five tests at family-wise $\alpha$=0.05;
Supplementary Note~3.3).

\subsection*{Enhanced gold standard}
\label{sec:enhanced_gold}

Error analysis (Results) showed the original annotation had omitted institutional
identifiers the LLMs correctly detected, penalizing correct redactions, deflating
F1 and obscuring differences between architectures. We therefore re-annotated a
subset for relative comparison. From the 414~candidate spans surfaced corpus-wide
we selected the 10~notes with the largest model--gold discrepancy; a domain expert
adjudicated 49~in-context instances covering 22~unique institutional terms,
confirming all 49 as TRUE\_PHI. Annotating those terms at every occurrence added 227~spans
(209~Geographic Data, 9~Name, 9~Other Unique~ID), for 1,758~total versus
1,531~original; institutional terms were assigned to Geographic Data, which is why
that category grows far beyond its corpus-wide count (Supplementary Note~2.5). Three
biases follow: the subset overrepresents institutional PHI density; re-annotation
covered only the 22~surfaced terms; and the candidates came from model output,
partly crediting those models for spans they surfaced. Expert adjudication
establishes the added spans are genuine PHI, not that they exhaust it, so the
enhanced standard serves only for relative comparison on these 10~notes. All
multi-stage architectures, and a five-trial re-evaluation of the three prompt
conditions, were scored against it.

\subsection*{Evaluation pipeline}
\label{sec:pipeline}

Each gold PHI span is a \textbf{true positive (TP)} when its text is absent from
the output, and a \textbf{false negative (FN)} when an exact substring search still
finds it, when masking is partial, or when no output is returned. A \textbf{false
positive (FP)} is an emitted placeholder matching no gold annotation. Matching is
type-agnostic, so a date masked as \phitag{NAME} still counts as TP. Recall is TP/(TP+FN) and precision TP/(TP+FP), whose denominator mixes gold spans
with emitted placeholders; that unit mismatch cuts both ways, raising precision
where a system merges adjacent gold spans and lowering it where one is split across
several. Text also occurring as legitimate non-PHI (``May'' as name vs.\ month) can
produce spurious counts. Outputs shorter than 50\% of the note are scored as
failures with all spans FN. False negatives were categorized as \textbf{canonical} or \textbf{institutionally
situated} PHI. Placeholder alignment, TiDE's surrogate-based precision scoring, the
mixed-unit bias, and the full rule set are in Supplementary Note~1.2. Figures use
Matplotlib in the soft-fill style of PubliPlots~\cite{botas2025publiplots}.

\subsection*{Ethical considerations}

This study was conducted under IRB protocol H-52222 at Baylor College of
Medicine / Texas Children's Hospital. All data remained within the institutional
environment, and LLM inference via AWS Bedrock ran under a HIPAA-compliant Business
Associate Agreement under which prompts and outputs are neither retained nor used
for training. LLMs are the object of study here; any use of AI tools in manuscript
preparation is disclosed in Additional Contributions.

\section*{Results}
\label{sec:results}

Recall is the primary safety metric, since a missed span is a potential privacy
violation whereas an over-redacted one only degrades data utility. We select on F1
to keep the precision cost of recall gains visible, but because F1 weights the two
errors equally we also tabulate the recall-weighted $F_2$ ($5PR/(4P+R)$;
Tables~S3 and~S8), which
reorders only adjacent pairs and leaves the leaders unchanged.

\subsection*{LLMs outperform traditional de-identification on pediatric oncology notes}
\label{sec:overall_results}

\begin{figure}[htbp]
\centering
\includegraphics[width=\textwidth]{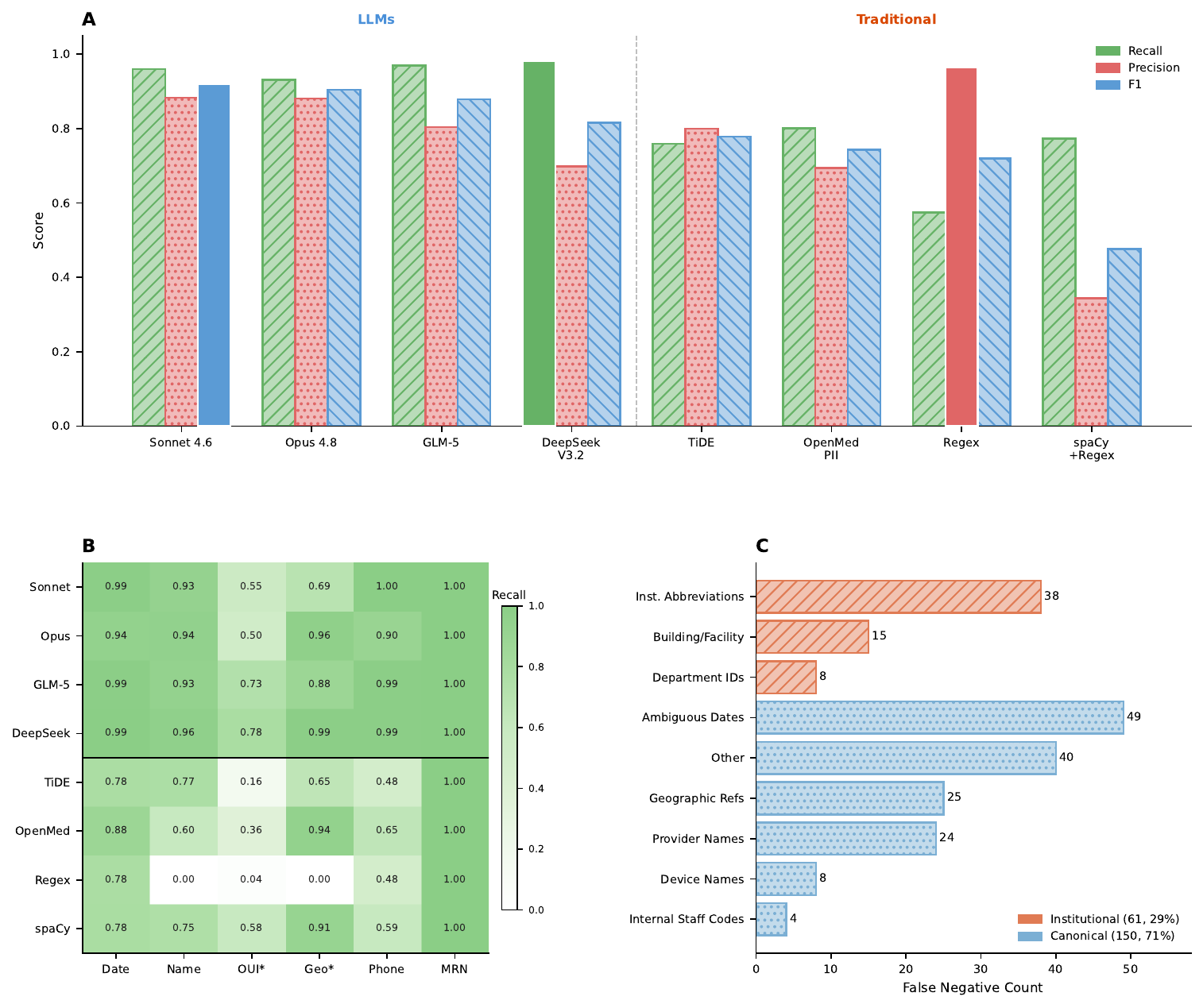}
\caption{LLMs outperform traditional de-identification approaches.
\textbf{(A)}~Recall, precision, and F1 for the top~4 LLMs and 4 traditional
baselines (\prm{Baseline} prompt, 100 notes, 5,322 spans; all 8~LLMs in
Table~S3 and Supplementary
Figure~1).
\textbf{(B)}~Per-category recall heatmap for the same systems, with categories
ordered by gold-standard frequency and systems by overall F1, reveals shared
weakness on the institutionally enriched categories (Other Unique~ID [OUI],
Geographic Data); the four categories with $\leq$5 gold spans are omitted here
and reported in Table~S4.
\textbf{(C)}~Sonnet~4.6 false negative decomposition on the primary \prm{Baseline}
run (Table~S3; 211~FN): 29\% are
institutionally situated PHI (61 of 211; institution abbreviations~38,
building/facility~15, department/clinic codes~8; see taxonomy in
Table~S7); the remaining 150 are canonical, of which the
internal staff codes (pager and Voalte extensions,~4) are shown separately
because Table~S7 lists them as borderline.}
\label{fig:figure2}
\end{figure}

Under identical input (\prm{Baseline} prompt for LLMs), LLMs substantially
outperformed all traditional approaches (Figure~\ref{fig:figure2}). The spaCy NER
+ regex baseline reached 77.3\% recall at only 34.4\% precision (7,832 false
positives), since general-purpose NER labels medications, diagnoses, and anatomy as
entities; regex alone reached 57.4\% recall at 96.4\% precision, working for
structured but not context-dependent PHI. Stanford TiDE reached 75.9\% recall
(F1=0.779) and OpenMed PII 80.1\% recall at 69.4\% precision (F1=0.743), the
highest non-LLM recall but lower F1 (Table~S3,
Supplementary Figure~1). The same ordering held on the 49-note
validation set (best LLM F1 0.894 vs.\ 0.866), though the margin narrowed on that
canonical-PHI-dominated corpus, and on $F_2$ the best-LLM advantage over TiDE
there falls to 0.001 (Supplementary Note~4).

Among LLMs, Sonnet~4.6 combined 96.0\% recall with 88.3\% precision for the best
F1 (0.920 in the primary run; 0.918$\pm$0.001 across five trials); Opus~4.8
matched its precision at lower recall (93.1\%, F1=0.905);
and the six LLMs without a failure mode exceeded TiDE's recall
by 0.15--0.22 (Table~S3), the exception being
GPT-oss-20B, whose truncation puts it below TiDE. The trade-off is
model-dependent: under \prm{Baseline} the highest-recall models over-redact enough
to generate thousands of false positives (DeepSeek~V3.2 2,264~FP at 98.3\% recall;
Kimi~K2.5 3,251~FP at 96.9\%), whereas Sonnet~4.6 produces only 679~FP at
comparable recall.

\subsection*{Per-category analysis reveals shared weaknesses on institutionally situated PHI}
\label{sec:TiDE_comparison}

Because aggregate recall is dominated by Date spans (72.1\% of gold annotations),
per-category results (Figure~\ref{fig:figure2}B; all twelve systems and ten
categories in Table~S4) show both \emph{why} LLMs
outperform traditional systems and \emph{where} they fail. TiDE achieves perfect
recall on MRN but fails where context is required: Other Unique ID (15.7\%), Phone
(47.9\%), and Geographic Data (65.0\%). Averaged over
the top~4 LLMs, recall exceeds TiDE by 0.49 on Phone and 0.48 on Other Unique ID
--- identifiers in non-standard formats such as pager codes and internal extensions
that evade regex but are readable from context (Supplementary Note~2.3). The panel
also reveals the LLMs' shared weakness: even the best models reach only 50--78\%
recall on Other Unique ID and 69--99\% on Geographic Data, the categories most
enriched for institutionally situated PHI.

\subsection*{In-context learning enables control over the precision--recall trade-off}
\label{sec:prompt_engineering}

Unlike TiDE or regex baselines, whose behavior is fixed by their pattern
libraries, LLMs can be steered by prompt design alone. We developed three prompt
conditions through iterative error analysis on Sonnet~4.6, then evaluated all
across the model set. Decomposing Sonnet~4.6's 211~\prm{Baseline} false negatives
(Figure~\ref{fig:figure2}C, a representative run bracketed by the five-trial means
in Table~S9) identified institutional PHI as the blind
spot: 29\% of misses (61/211), dominated by institution abbreviations (38) and
building or facility names (15).
Table~S7 gives the subcategory taxonomy with failure mechanisms
and recovery rates.

Appending those four categories to the baseline instructions (\prm{Targeted})
reduced Sonnet~4.6's institutional false negatives from 61 to 13
(\textbf{78.7\% recovery}), with the largest gains on Other Unique ID (recall
0.548 to 0.791) and Name (0.928 to 0.971), raising overall recall from 0.958 to
0.975 (5-trial means; Table~S9). It also carried a
precision penalty (0.881 to 0.807), as the added instructions over-redacted
clinical content in six recurring categories (Supplementary Note~2.2).
\prm{Precision} keeps those categories and appends anti-over-redaction
WRONG~$\rightarrow$~RIGHT pairs, recovering precision to 0.829 for a modest recall
cost (0.975 to 0.969) and F1=0.893$\pm$0.004; the full progression nets higher
recall (+0.011) at lower precision ($-$0.052).

The trend holds across the full model set (Figure~\ref{fig:figure3}A--B;
Tables~S5 and~S6;
per-category recall in Supplementary
Figures~2 and~3),
with \prm{Targeted} improving recall (Opus~4.8 +0.023, Kimi~K2.5 +0.018,
Sonnet~4.6 +0.014) and \prm{Precision} improving precision for all 7~models with
valid output (mean +0.054). All three prompts carry worked examples; only those
naming content to \emph{preserve} recover the clinical text category-only
instructions over-redact~\cite{iscience2025benchmark}. Magnitude varies by model:
those with severe baseline precision deficits benefit most (GPT-oss-120B P=0.198 to
0.924; Kimi 0.613 to 0.816; DeepSeek 0.698 to 0.806), whereas MiniMax regresses on
recall under \prm{Targeted} ($-$0.141) and Sonnet~4.6 scores its highest F1 under
\prm{Baseline}, whose high precision the instructions can only cost; the optimal
prompt is therefore model-dependent. That F1 penalty must be read with caution:
as the next section shows, many of \prm{Precision}'s additional ``false
positives'' are correct redactions of institutional PHI the original standard
failed to annotate.

\subsection*{Error analysis reveals annotation gaps in the gold standard}
\label{sec:annotation_gap}

While comparing multi-agent pipelines against our best single-pass model, we
noticed they flagged institutional PHI absent from the gold standard. Rather than
count these as false positives, we asked whether the standard itself was
incomplete. Manual inspection confirmed annotators had systematically missed
institution-specific identifiers: operating room location codes (CC~OR, LT~OR,
MW~OR, WC~OR, WT~MAIN~OR, GIPS), hospital abbreviations (TCH, TXCH, BCM), and
campus or building names (Mark Wallace Tower, West Campus), all PHI under the
catch-all and expert-determination provisions. The scarcity of reliable annotations
motivates label-free evaluation~\cite{team_phi_2025}; our results show they are
also \emph{incomplete}. Expert re-annotation of the 10~highest-discrepancy notes
confirmed all 49~adjudicated instances as PHI and added 227~spans; the protocol and
its biases are in Methods.

\textbf{Re-evaluation against enhanced gold.} On the enhanced 10-note subset
(1,758~spans, 5~trials per prompt; Figure~\ref{fig:figure4}C,
Table~S10), annotation gaps
disproportionately penalize institution-aware prompts. \prm{Baseline} recall is
0.847$\pm$0.002 here versus 0.958$\pm$0.002 on the original 100-note standard,
because it leaves the newly-annotated institutional terms unredacted. Those two
numbers differ in note set as well as standard, so we held the notes fixed: against
the \emph{original} annotation of these same 10~notes, \prm{Baseline} recall is
0.972$\pm$0.003, so re-annotation accounts for $-$0.125 of the $-$0.111 net change
and note selection for $+$0.014 (Supplementary Note~2.6). \prm{Targeted} recovers recall to
0.980$\pm$0.000 at precision 0.810$\pm$0.004 (F1=0.887$\pm$0.002), and
\prm{Precision} holds recall (0.981$\pm$0.000) while recovering precision to
0.844$\pm$0.003 (F1=0.907$\pm$0.002) --- a \prm{Baseline}-to-\prm{Targeted} recall
gap far larger on the enhanced gold (0.133) than on the original (0.017).

\subsection*{Multi-agent architectures confirm single-pass sufficiency}
\label{sec:multi_agent_results}

None of the 14~multi-agent and ensemble configurations (Supplementary Note~3,
Tables~S11 and~S12) improved F1 over
the single-pass \prm{Precision} prompt. The three reproducibility-tested methods
(5~trials each, Table~S13) reached comparable mean F1
(0.906--0.908) with overlapping 95\% trial-resampled CIs
(Figure~\ref{fig:figure4}A--B), though our trial count cannot establish formal
equivalence. The agentic pipeline's intra-method variability
(SD$_{\text{F1}}$=0.018) exceeds inter-method differences, and 4~of its 5~trials
fell below single-pass (median~0.898 vs.\ mean~0.906, pulled up by one
high-precision run at F1=0.942), so a typical agentic run underperforms. Ensemble
voting reaches marginally higher recall (Cross-Model Vote 0.986 vs.\ 0.981) at a
precision cost and 6$\times$ the inference cost. Across the four top LLMs, 20.3\%
of false negatives are shared, dominated by ambiguous partial dates and
institutional identifiers (Supplementary Note~2.4).

\section*{Discussion}
\label{sec:discussion}

Our results establish three findings. First, LLMs substantially outperform
purpose-built systems, with the advantage concentrated in categories requiring
contextual reasoning. Second, in-context learning enables control over the
precision--recall trade-off, and which prompt looks best depends on how completely
the standard annotates institutional PHI: \prm{Baseline} wins on the original
annotations and loses on the corrected ones, not because \prm{Precision} improves
but because \prm{Baseline}'s recall collapses once the institutional terms it
leaves unredacted are counted.

Third, no multi-agent architecture improved F1 over single-pass. Those
configurations proved useful instead as a \emph{discovery tool} for enumerating
annotation gaps at scale, though not uniquely so, since the single-pass
\prm{Precision} prompt flagged the same identifiers. The prover-verifier
framework~\cite{openai2024proververifier} that motivated this exploration yields
no gain here because the bottleneck is specification --- which institutional terms
to redact --- not verification of a given redaction. Recommended deployment
configurations are given in Supplementary Note~3.5.

\subsection*{Limitations}

First, our corpus (100~notes, single institution) limits generalizability. The
49-note USDHUB validation set (Supplementary Note~4) confirms the LLM F1 advantage
persists but narrows on canonical-PHI-dominated corpora; because USDHUB is from the
same institution, it establishes robustness across specialties, note types, and
annotation methods, but not across institutions. Cross-institutional generalization
and the transferability of our site-specific addenda remain to be established. Second, the 10~re-annotated notes were selected by
highest model--gold discrepancy, biasing toward vindicating the model, so
enhanced-gold results are relative comparisons across prompts, not corpus-wide
estimates. Third, TiDE was run unprovisioned on the primary corpus; on USDHUB both
extremes were near-identical (recall 0.976 vs.\ 0.971; Supplementary Note~4.1), so
the LLM advantage holds whether or not pattern-based systems receive site-specific
information. Fourth, the \prm{Targeted} and \prm{Precision} addenda came from \prm{Baseline}
error analysis on this same corpus, so their gains are in-sample. Fifth, we report no subgroup or fairness analysis: the attributes that would define
subgroups are themselves the PHI under removal and were never extracted, so whether
recall differs across patient groups is unresolved. Sixth, both GPT-oss models
showed failure modes unrelated to de-identification capability (GPT-oss-20B output
truncation, depressing recall; GPT-oss-120B unstable redaction at precision 0.198),
unlike the other six (Table~S5).

Finally, that multi-agent architectures fail to beat single-pass likely reflects
the task itself: de-identification is a single-read problem in which all necessary
context is already in the note. Additional passes let the model second-guess correct
decisions, and coordination adds noise without information (Supplementary
Note~3.4). This matches findings outside the clinical
domain, where multi-agent gains are often minimal and failures trace to
specification rather than model
capability~\cite{cemri2025multiagentllmsystemsfail}, and where a single agent with
strong prompts matches multi-agent discussion, the latter winning only where the
prompt leaves the task
underspecified~\cite{wang2024rethinkingboundsllmreasoning}; our \prm{Targeted} and
\prm{Precision} prompts supply that specification.

\subsection*{Future directions}

Promising directions include knowledge distillation, using high-performing LLMs as
teacher labelers for small, locally-deployable models at 100$\times$ lower
cost~\cite{kim2024generalizing,shield2026,zambare2026fair}; adversarial
verification against an LLM attempting
re-identification~\cite{diri2024,carrell2013hiding,carrell2020resilience};
multi-institutional validation; and gold standard refinement to include
institutionally situated PHI.

\section*{Conclusion}
\label{sec:conclusion}

On 100 pediatric oncology notes, LLMs beat purpose-built de-identification on
recall --- its core promise --- by 0.20 over Stanford TiDE, and the gap is widest
exactly where pattern matching cannot reach: identifiers whose PHI status depends
on institutional context.

Prompting, not retraining, is the adaptation mechanism. Naming the institutional
categories a HIPAA-aligned prompt misses recovers 79\% (48/61) of them, and
anti-over-redaction instructions then restore precision, within a single LLM call
and impossible with purpose-built systems short of retraining. Multi-agent
architectures do not help; none of 14~configurations improved F1 over a single pass
with the same prompt, because de-identification is a single-read task in which the
second pass has no information the first lacked. Ensembles buy marginally higher
recall at a precision cost and 3~calls per note. And LLMs proved good enough to
audit their own reference standard: what looked like over-redaction was largely PHI
the annotators had missed, expert adjudication confirming all 49~in-context
instances as true PHI.

Three consequences follow. De-identification should be evaluated per category, not
on aggregate metrics that hide institutional PHI; gold standards deserve auditing
before they are trusted as ground truth; and effort belongs in institution-specific
prompts rather than additional passes. LLMs cost more per note, but that cost buys
adaptation without retraining and a check on the standard itself. Our evaluation
framework is released open-source.


\section*{Funding}
This research was supported by a fellowship from the Gulf Coast Consortia on
the NLM Training Program in Biomedical Informatics and Data Science
(T15~LM007093); the National Science Foundation Graduate Research
Fellowship Program (NSF GRFP Fellow ID 2024370642); the Fund for
Innovation in Cancer Informatics; the Cancer Prevention and Research
Institute of Texas (CPRIT, RP240131); the Chan Zuckerberg Initiative
(2023-332162); the National Institutes of Health (NIH, U54NS093793 and
OT2OD040565); the Eunice Kennedy Shriver National Institute of Child Health
and Human Development of the NIH (P50HD103555); the Chao Endowment; the
Huffington Foundation; and the Jan and Dan Duncan Neurological Research
Institute at Texas Children's Hospital.

\section*{Additional Contributions}
We thank the Texas Children's Hospital Office of Research Data, which
provided the independently de-identified USDHUB note set used for our
within-institution validation (Supplementary Note~4).
No AI-assisted tools were used for study design, analysis, or primary manuscript
drafting. Generative AI tools were used only for proofreading and
typographical/grammatical correction of author-written text; they were not used
to generate scientific content, analyze data, or draft substantive passages of
the manuscript. In accordance with COPE's position and JAMIA policy, no AI or
NLP tool is listed as an author; the authors reviewed and verified all text and
take full responsibility for the integrity, accuracy, and originality of all
content.

\section*{Conflicts of Interest}
The authors declare no competing interests.

\section*{Data Availability}
The evaluation pipeline and code are openly available at
\url{https://github.com/LiuzLab/phi-scrubber-evaluation}. The underlying
clinical notes cannot be shared because they are protected patient health
information governed by IRB protocol H-52222 and institutional/HIPAA data-use
restrictions; they are not available for public deposition or on request.
De-identified aggregate metrics that support the findings of this study are
provided within the article and its supplementary materials.

\section*{Author Contributions}
M.B.N. and D.P. contributed equally as co-first authors: conceptualization,
methodology, software, formal analysis, investigation, visualization, and writing.
J.P.W., C.L., A.A.O., and S.D. contributed to data curation and investigation
(PHI annotation and re-annotation). M.Z. provided clinical expertise, adjudication
of institutional PHI, annotation, and review and editing of the writing. Z.L. is
the principal investigator and contributed supervision and funding acquisition.
H.H.J. contributed supervision, project administration, and review and editing of
the writing. All authors reviewed and approved the final manuscript.

\bibliography{references}

\clearpage
\section*{Figures}

\noindent\textbf{Alt text:} Three-panel study-design figure. Panel A is a
schematic showing eight large language models and traditional baselines being
evaluated on 100 pediatric oncology notes containing 5,322 PHI spans under
three prompt conditions. Panel B shows a synthetic clinical note with canonical
HIPAA identifiers highlighted in blue and institutionally situated identifiers
(facility abbreviations, building names) highlighted in orange. Panel C is a
bar or pie chart of the original gold-standard PHI category distribution over
the six well-populated categories, dominated by Date (72.1 percent) and Name
(22.3 percent), with institutionally enriched categories comprising 3.7 percent
of spans, a lower bound given the under-annotation described in Results.

\noindent\textbf{Alt text:} Three-panel performance-comparison figure. Panel A
is a grouped bar chart of recall, precision, and F1 for the top four LLMs and
four traditional baselines under the Baseline prompt, showing LLMs above
traditional systems on F1. Panel B is a per-category recall heatmap in which
the Other Unique ID and Geographic Data columns are lightest (lowest recall)
across systems, indicating shared weakness on institutionally situated PHI.
Panel C is a horizontal bar chart breaking down Sonnet~4.6's 211 false
negatives, with 29 percent (61 spans) attributed to institutionally situated
PHI and the remaining 150 to canonical categories.

\begin{figure}[htbp]
\centering
\includegraphics[width=\textwidth]{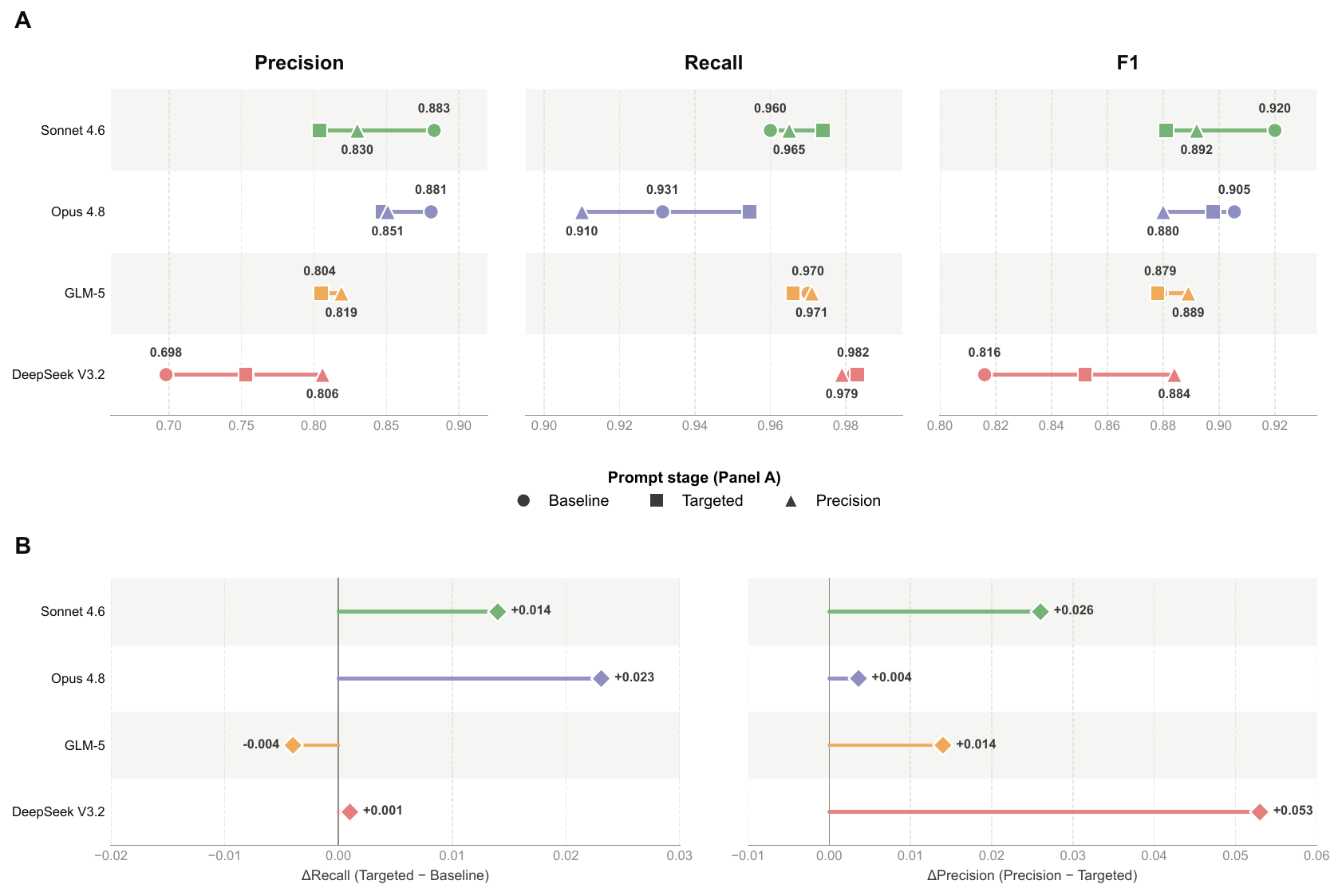}
\caption{In-context learning enables control over the precision--recall trade-off.
\textbf{(A)}~Precision, recall, and F1 for the top~4 LLMs across the three
prompt versions (\prm{Baseline}~$\bullet$, \prm{Targeted}~$\blacksquare$,
\prm{Precision}~$\blacktriangle$; single run per model, all~8 in Table~S5). Each
horizontal line spans the minimum-to-maximum of a model's three stage values,
so a non-monotonic \prm{Targeted} stage visibly stretches the line rather than being
hidden; \prm{Baseline} and \prm{Precision} values are labeled. Effect magnitude and
direction are model-dependent.
\textbf{(B)}~Problem-specific deltas (single run), same model order and colors
as~(A), isolating what each prompt revision buys: $\Delta$Recall(\prm{Targeted}$-$\prm{Baseline})
is the recall gain from institutional targeting, and
$\Delta$Precision(\prm{Precision}$-$\prm{Targeted}) is the precision recovery from the
anti-over-redaction instructions. The \prm{Precision} prompt lifts precision for
\emph{all} models (cross-model mean over the 7~models with valid output:
+0.054) and acts as a corrective: gains are largest for models that over-redact
under the baseline prompt (GPT-oss-120B~$0.198\rightarrow0.924$,
Kimi~$0.613\rightarrow0.816$, DeepSeek~$0.698\rightarrow0.806$ across the full
\prm{Baseline}$\rightarrow$\prm{Precision} progression; full values in
Tables~S5 and~S6) and
near-zero for models whose native precision is already high. Sonnet~4.6
reproducibility (5~trials per stage on the full 100-note corpus; SD~$\leq$0.007
at every stage) confirms these are systematic prompt effects, not run-to-run
noise (Table~S9).}
\label{fig:figure3}
\end{figure}

\noindent\textbf{Alt text:} Two-panel figure on prompt effects. Panel A plots
precision, recall, and F1 for the top four LLMs across three prompt versions
(Baseline, Targeted, Precision) as horizontal min-to-max ranges per model,
showing model-dependent magnitude and direction. Panel B is a bar chart of
problem-specific deltas: the recall gain from institutional targeting and the
precision recovery from anti-over-redaction instructions, with the largest
precision gains for models that over-redact under the baseline prompt and
near-zero gains for models with already-high native precision.

\begin{figure}[htbp]
\centering
\includegraphics[width=\textwidth]{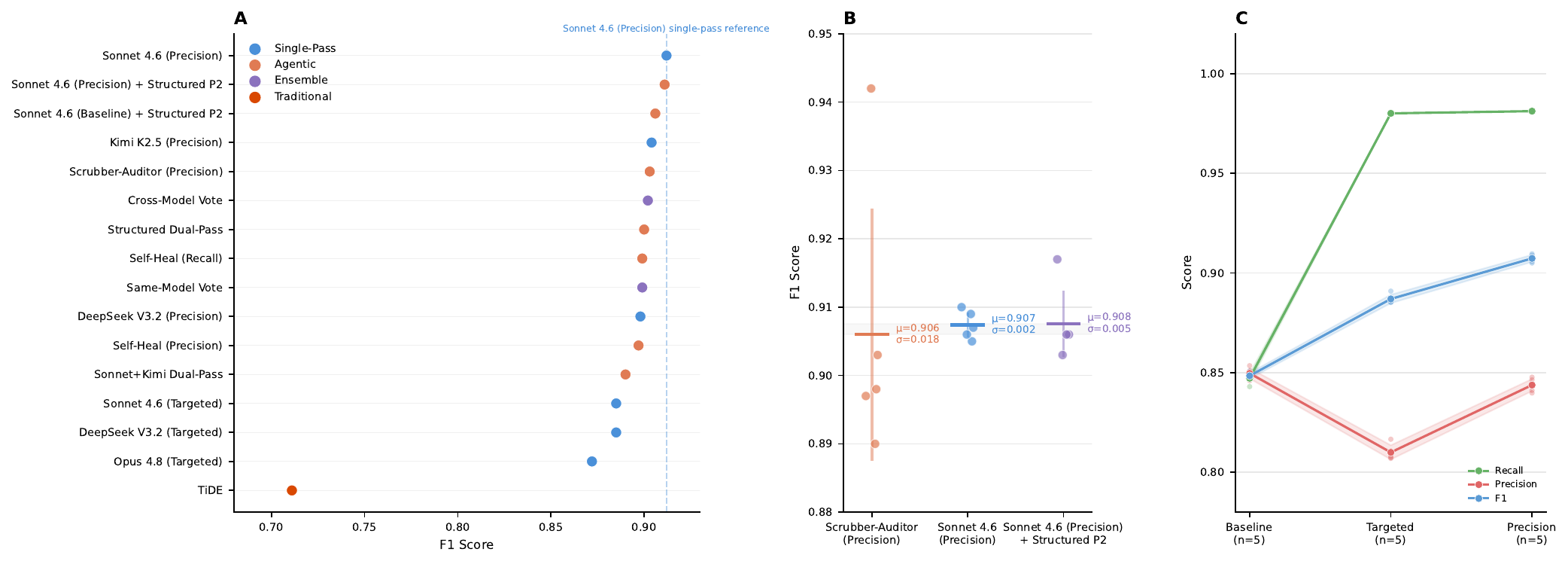}
\caption{Enhanced gold standard validation and agentic architecture comparison.
\textbf{(A)}~Configuration landscape: F1 for the 16~configurations that the panel
displays, colour-coded by class (single-pass, agentic, ensemble, traditional) and
ranked on the enhanced gold standard; 10~further configurations are omitted for
legibility and appear in Table~S12, which lists all 26. The
omitted set is the weakest-performing tail, except that TiDE is retained as the
traditional-system reference. Single-pass
Sonnet~(\prm{Precision}) (dashed line) matches or exceeds every multi-agent
variant, and self-refinement and ensemble voting fail to beat it.
\textbf{(B)}~Reproducibility: 5-trial strip plots for the top 3 methods.
All three achieve closely comparable mean~F1 (0.906--0.908) with overlapping 95\% trial-resampled CIs,
but the Scrubber--Auditor (\prm{Precision}) pipeline on Opus shows high variance
(SD$_{\text{F1}}$=0.018) vs.\ near-deterministic single-pass Sonnet
(SD$_{\text{F1}}$=0.002).
\textbf{(C)}~Sonnet~4.6 prompt progression on the enhanced gold standard
(10-note subset, 1,758~spans; 5~independent trials per version). Lines show
mean; shading shows $\pm$SD; points show individual trials. \prm{Baseline} recall is 0.847 here versus 0.958 on the
original 100-note gold standard (a different note set \emph{and} a different gold
standard) because it misses newly-annotated institutional terms; \prm{Targeted} recovers recall to 0.980; \prm{Precision} maintains recall while
improving precision (F1=0.907$\pm$0.002).
\textit{Terminology:} \emph{single-pass} = one LLM call per note;
\emph{dual-pass} = two sequential LLM calls; \emph{structured pass~2} = a second
pass that ingests the first pass's output in a structured format;
\emph{Scrubber--Auditor} = an agentic pipeline in which a generator LLM
(scrubber) is checked by a second auditor LLM. \prm{Baseline}, \prm{Targeted}, and \prm{Precision}
denote the three prompt versions.}
\label{fig:figure4}
\end{figure}

\noindent\textbf{Alt text:} Three-panel figure on agentic architectures. Panel
A ranks 16 configurations, spanning single-pass, agentic, ensemble, and
traditional approaches, by F1 on the enhanced gold standard, with a dashed line
marking single-pass Sonnet (Precision) at or above all multi-agent variants. Panel B shows 5-trial strip plots for the top three
methods with closely overlapping mean F1 of 0.906 to 0.908, but visibly wider
scatter for the Opus Scrubber--Auditor pipeline than for near-deterministic
single-pass Sonnet. Panel C plots Sonnet 4.6 recall, precision, and F1 across
the Baseline, Targeted, and Precision prompts on the enhanced gold standard,
showing Baseline recall dropping to 0.847, Targeted recovering recall to 0.980,
and Precision maintaining recall while improving precision to F1 = 0.907.


\end{document}


\appendix
\renewcommand{\thetable}{E\arabic{table}}
\setcounter{table}{0}
\renewcommand{\thesuppfigure}{E\arabic{suppfigure}}
\setcounter{suppfigure}{0}

\section*{Extended Supplementary Material}

\section*{Extended Limitations and Future Directions}

The following points were raised as potential extensions to the primary
analysis. Each would require additional experimentation, re-scoring, or
external data beyond the scope of the present study, and we therefore document
them here as explicit limitations and directions for future work rather than
as new results.

\subsection*{Robustness to paraphrase and residual re-identification risk}

Our evaluation measures whether direct and institutionally situated
identifiers are removed (recall) and whether clinical content is preserved
(precision). It does not measure resistance to \emph{re-identification by
inference}, for example an adversary reconstructing a patient's identity
from a constellation of residual quasi-identifiers (rare diagnoses, unusual
treatment timelines, or distinctive narrative phrasing) even after all
HIPAA-enumerated identifiers are redacted. Nor does it test whether an LLM
adversary could re-identify individuals by paraphrasing or cross-referencing
de-identified notes against external corpora. Prior work on ``hiding in plain
sight'' surrogate replacement~\cite{carrell2013hiding,carrell2020resilience}
and adversarial re-identification~\cite{diri2024} provides a framework for
this analysis; applying such an adversarial protocol to our LLM outputs is an
important direction for future work.

\subsection*{Macro-averaged and note-level metric aggregation}

The precision, recall, and F1 values reported in the main text are
micro-averaged over all gold spans, so frequent categories (e.g., Name, Date)
dominate the aggregate. Macro-averaging across PHI categories, or reporting
note-level distributions of these metrics, would give equal weight to rare but
high-risk categories and would better characterize worst-case behavior on
individual notes. The per-category recall heatmaps and per-note recall
distributions (Supplementary Figures) provide a partial view, but a full
macro-averaged re-tabulation of all systems and prompt versions is deferred to
future work.

\subsection*{Offset-based span scoring}

Our scoring aligns system output against the gold standard at the token/word
level (via text alignment), counting a span as detected when the identifying
text is redacted. A stricter \emph{offset-based} scoring scheme, requiring
exact character-boundary agreement between predicted and gold spans, would
penalize partial-boundary matches (e.g., redacting ``Children's Hospital'' when
the gold span is ``Texas Children's Hospital'') that our current scheme may
credit. Because boundary conventions differ across the LLM (free-text
placeholder) and traditional (offset-emitting) systems, a fully offset-based
comparison would require re-normalizing all outputs to a common span
representation. We expect the qualitative ranking to be robust to this choice
but have not quantified it.

\subsection*{Model and API version identifiers}

The eight LLMs are identified in the main text by their public model names and
sizes. Exact provider model identifiers govern reproducibility, since hosted
models can be updated silently. Table~\ref{tab:model_ids} gives the exact
identifier used to invoke each model.

\begin{table}[H]
\centering
\caption{Exact AWS Bedrock model identifiers for the eight benchmarked LLMs.
All models were invoked through the Bedrock Converse API in region
\texttt{us-east-1} at \texttt{temperature=0.0} (except Opus~4.8, which rejects
the parameter), \texttt{max\_tokens=65{,}000}, with up to 3 retries under
exponential backoff.}
\label{tab:model_ids}
\small
\begin{tabular}{lll}
\toprule
\textbf{Name in text} & \textbf{Bedrock model identifier} & \textbf{Provider} \\
\midrule
Claude Opus 4.8   & \texttt{us.anthropic.claude-opus-4-8}$^\dagger$   & Anthropic \\
Claude Sonnet 4.6 & \texttt{us.anthropic.claude-sonnet-4-6}$^\dagger$ & Anthropic \\
GPT-oss-120B      & \texttt{openai.gpt-oss-120b-1:0}                & OpenAI \\
GPT-oss-20B       & \texttt{openai.gpt-oss-20b-1:0}                 & OpenAI \\
GLM-5             & \texttt{zai.glm-5}                             & Zhipu AI \\
Kimi K2.5         & \texttt{moonshotai.kimi-k2.5}                   & Moonshot \\
MiniMax M2.5      & \texttt{minimax.minimax-m2.5}                   & MiniMax \\
DeepSeek V3.2     & \texttt{deepseek.v3.2}                          & DeepSeek \\
\bottomrule
\multicolumn{3}{l}{\small $^\dagger$Cross-region inference profile identifier (the \texttt{us.} prefix). On-demand invocation of} \\
\multicolumn{3}{l}{\small \quad the bare foundation-model identifier fails with a \texttt{ValidationException} for these models.} \\
\multicolumn{3}{l}{\small All eight were verified to accept the full 65{,}000-token output ceiling, required so that the longest} \\
\multicolumn{3}{l}{\small \quad notes ($\sim$65.8k characters) can be reproduced without truncation.} \\
\end{tabular}
\end{table}

\noindent\textbf{Inference dates.} All LLM inference reported in this study was
executed between 2026-05-30 and 2026-06-28: the primary-corpus runs across all
eight models and three prompt conditions between 2026-05-30 and 2026-06-04, the
Sonnet~4.6 reproducibility trials through 2026-06-28, and the USDHUB validation
runs on 2026-06-25 and 2026-06-26. These bounds are derived from the timestamps
of the stored run artifacts; per-invocation timestamps were not recorded by the
evaluation harness, so we report the execution window rather than exact
per-model inference times. Because Bedrock model endpoints can be updated
without a change to the identifier, results should be interpreted as
characterizing these endpoints as served during that window.

\subsection*{John Snow Labs Spark NLP baseline}

We were unable to include John Snow Labs' Spark NLP for Healthcare
de-identification pipeline, a widely used commercial baseline, because it was
unavailable to us under its licensing terms during the study period. Its
omission means our non-LLM comparison set (TiDE, OpenMed PII, spaCy+regex,
regex-only) does not include a fully-provisioned commercial system, which may
understate the strongest achievable traditional-pipeline performance.
Benchmarking against Spark NLP is left for future work.

\clearpage

\section*{Extended Detailed Results and Methods}

This section collects the detailed tables, full prompt text, error case studies,
reproducibility analyses, and cost/latency breakdowns that support the headline
results summarized in the main JAMIA supplement (Supplement~1). Nothing here is
new work; it is the complete detail behind the summarized findings, relocated so
that Supplement~1 remains focused on the key tables and figures a reader needs to
follow the main text.

\subsection*{Section E2: Full Prompt Text}

\subsubsection*{E2.1: Baseline Prompt (de-identification-only mode)}

\textbf{System prompt:}
\begin{quote}\small\ttfamily
You are a medical text deidentification expert. Your task is to return the
EXACT same text as provided, with PHI (Protected Health Information) replaced
by generic placeholders. DO NOT change the structure, formatting, or wording
of the text in any way. Simply replace identifiers with placeholders.
CRITICAL: Return the text EXACTLY as given, with ONLY the PHI replaced by
placeholders.
\end{quote}

\textbf{User prompt} enumerates the 18 HIPAA Safe Harbor identifier categories
with examples: "Shelby Smith is a 3yo F" becomes "[NAME] is a 3yo F";
"She was born on 3/1/11" becomes "She was born on [DATE]".

\subsubsection*{E2.2: Targeted Prompt (adds institutional PHI)}

Appends to Baseline a block titled "ADDITIONAL REQUIRED REDACTIONS" with four
categories: (A)~physician/staff names adjacent to credentials, (B)~medical-staff
pager and Voalte numbers, (C)~institution names and abbreviations (TCH, TXCH,
BCM), and (D)~building/facility names. Each includes
WRONG~$\rightarrow$~RIGHT example pairs.

\subsubsection*{E2.3: Precision Prompt (adds anti-over-redaction)}

Retains all Targeted content and appends a "DO NOT OVER-REDACT" block targeting
six false-positive categories with WRONG~$\rightarrow$~RIGHT examples:
(1)~clock times, (2)~lab/vital values, (3)~medication doses, (4)~ages under~90,
(5)~relationship words, (6)~clinical abbreviations. Closes with a safety caveat:
"when genuinely ambiguous and could identify a specific person, prefer to
redact."

\subsubsection*{E2.4: Scrubber--Auditor Pipeline Prompts}

\textbf{Stage 1: Sonnet Scrubber (recall-maximizing):}
\begin{quote}\small\ttfamily
You are a specialized PHI scrubbing engine. [...] BE THOROUGH. It is critical
that NO PHI leaks through. When in doubt, REDACT. [...] ALSO REDACT these
commonly missed institutional identifiers: hospital names AND abbreviations,
NAMED building/facility names, device/DME company names, pharmacy names,
staff pager numbers. [...] DO NOT REDACT: clock times, generic location
descriptors (OR, PACU, ICU, Pod A), clinical department TYPES without
institution name, procedure codes, medication names, lab values, vital signs.
\end{quote}

\textbf{Stage 2: Opus Precision Auditor:}
\begin{quote}\small\ttfamily
You are a PHI precision auditor. [...] RESTORATION RULES: restore a
[PLACEHOLDER] back to its original value if the original value is OBVIOUSLY:
a medication name/dosage, lab result, diagnosis, anatomical term, medical
abbreviation, vital sign, procedure name, clock TIME, or generic department
name. [...] NEVER RESTORE: calendar DATES, person NAMES, SPECIFIC institution
names (TCH, BCM, Baylor), SPECIFIC building names (Mark Wallace Tower, West
Campus), phone/pager numbers, device BRAND names, ANYTHING uncertain.
\end{quote}

\textbf{Auditor execution.} The Precision Auditor runs once per note over the
\emph{entire} scrubbed document, not independently per placeholder. The auditor
prompt receives the full original note (as PHI-containing reference) and the
full scrubbed note in a single call, and returns the complete precision-audited
note; restorations are applied wherever the model judges a placeholder's
original value to be obviously clinical content under the rules above.

\textbf{Prompt-scope caveat.} The scrubber prompt used in the pipeline
experiments retained a device-name clause from an earlier prompt iteration.
Device identifiers were not a PHI category in this corpus, so any device-name
redactions it produced could only be scored as false positives; the clause is
reported here for exact reproducibility.

Full prompt text available in the code repository:
\texttt{src/phi\_benchmark/prompts/} (Baseline, Targeted, Precision) and
\texttt{src/phi\_benchmark/pipelines/prompts.py} (pipeline stages).

\subsection*{Section E3: Error Case Studies}

These case studies expand on the error taxonomy summarized in Supplement~1
(institutionally situated PHI taxonomy), providing the verbatim note excerpt
behind the institutional-abbreviation example.

\textbf{Case study 1: institutional abbreviation recovery.}
The abbreviation "TCH" appeared as a false negative for most LLMs under
Baseline. In context:

\begin{quote}
\small\texttt{Patient was seen at \textbf{TCH} Main Campus Hematology Center...}
\end{quote}

Under Baseline, models treated "TCH" as a clinical abbreviation. Under Targeted, Sonnet
4.6 recovered 48/61 institutional PHI misses (across all subcategories). The 13 remaining
occurred in complex compound expressions where institutional names were
embedded within clinical descriptors.

\subsection*{Section E4: Multi-Agent Reproducibility, Self-Refinement, Failure Analysis, and Cost}

These analyses expand on the multi-agent workflow experiments summarized in
Supplement~1 (architecture summary and all-configurations results tables). The
headline conclusion, that no agentic configuration outperformed calibrated
single-pass de-identification, is stated in Supplement~1; the reproducibility
trials, statistical tests, failure mechanisms, self-refinement experiments, and
cost/latency accounting behind that conclusion are given here.

\subsubsection*{E4.1: Reproducibility and Run-to-Run Variability}

To quantify run-to-run variability we conducted 5~independent trials of
the three highest-F1 configurations from the all-configurations table
(Supplement~1): the agentic
Scrubber--Auditor~(Precision) (Opus), Sonnet~4.6 single-pass (Precision~prompt), and
Sonnet~(Precision) + Structured~Pass~2. Each trial was run independently against
the enhanced gold standard (10-note subset, 1,758~spans) using identical
prompts and model parameters (temperature$=$0).

\paragraph{Sources of non-determinism.} All models are called via the AWS
Bedrock Converse API with \texttt{temperature=0.0} and identical inputs
across trials. However, \texttt{temperature=0} does not guarantee
bitwise-identical outputs from cloud-served LLMs due to GPU floating-point
non-determinism (non-associative parallel arithmetic), infrastructure
routing across heterogeneous hardware, and cascade amplification (a single
token flip early in a 20--33K character output changes the entire downstream
sequence). Additionally, Claude Opus~4.8 (used by the Scrubber--Auditor
pipeline) rejects the \texttt{temperature} parameter via Bedrock
("temperature is deprecated for this model"), so greedy decoding cannot be
explicitly enforced for this model. This partly explains why the Opus-based
agentic pipeline exhibits higher variance (SD$_{\text{F1}}$=0.018) than the
Sonnet-based single-pass (SD$_{\text{F1}}$=0.002). No \texttt{seed}
parameter is available via the Bedrock Converse API; the observed
variability is entirely attributable to infrastructure-level
non-determinism, not any parameter under experimenter control.

\noindent The per-trial run-to-run variability table for the top-3 methods (5~trials each, enhanced gold standard) is given in Supplement~1, Supplementary Note~3.3; it is not duplicated here.

\paragraph{Variance profiles.} The headline reproducibility statistics (mean
F1, bootstrap CIs, and the significance tests) are reported in Supplement~1,
Supplementary Note~3.3. Beyond those, the three methods differ markedly in
\emph{where} their variance originates:

\begin{itemize}
\item \textbf{Single-pass Sonnet~(Precision)} is near-deterministic
  (SD$_{\text{F1}}$=0.002, range 0.905--0.910) with stable recall
  (0.981) and precision (0.840--0.848).
\item \textbf{Sonnet~(Precision) + Structured~P2} adds marginal recall
  (mean~0.982 vs 0.981) with low additional variance
  (SD$_{\text{F1}}$=0.005).
\item \textbf{Scrubber--Auditor~(Precision)} exhibits high variance
  (SD$_{\text{F1}}$=0.018, range 0.890--0.942) driven almost entirely
  by precision instability (0.836--0.935). When the Precision~Auditor
  aggressively restores over-redactions (Trial~2, FP=116), F1 reaches
  0.942; when it is conservative (Trial~4, FP=327), F1 drops to 0.890.
\end{itemize}

The agentic pipeline's best runs (F1=0.942) substantially exceed any single-pass
result, motivating the self-refinement strategies evaluated in Section~E4.3.

\noindent The per-trial prompt-version reproducibility table on the enhanced gold standard is given in Supplement~1, Supplementary Note~3.3; it is not duplicated here.

\noindent The per-trial full-corpus prompt-version reproducibility table (100~notes, 5,322~gold spans) is given in Supplement~1, Supplementary Note~3.3; it is not duplicated here.

\paragraph{Residual false positives on enhanced gold.}
Even under the enhanced gold, Sonnet~(Precision) reports 310--329~FP across 5~trials
(mean precision=0.844$\pm$0.003; per-trial values in Supplement~1,
Supplementary Note~3.3). Informal review of a sample
suggests that many of these represent additional annotation gaps (e.g.,
dates in vitals sections and follow-up appointments not marked in the gold
standard), but we have not performed systematic categorization or a second
re-annotation pass. The residual FP count should therefore be interpreted
as an upper bound on true over-redaction.

\subsubsection*{E4.2: Failure Analysis}

Why a second pass adds little (the Precision prompt already captures 222/227
institutional spans; the structured Pass~2 adds only 20 replacements; the
dedicated auditor restores zero) and how the same pipeline surfaced the
414~candidates behind the enhanced gold standard are summarized in Supplement~1,
Supplementary Note~3.4. The mechanism behind free-form dual-pass instability,
detailed next, is specific to this section.

\paragraph{Dual-pass variability and rewriting artifacts.}
The standard dual-pass architecture (Baseline Pass~1 + institutional recall
Pass~2) showed high run-to-run variability across 3~independent runs on
the original 100-note gold standard (5,322~spans): FP counts ranged from
532 to 1,524 (mean F1=0.904$\pm$0.043 vs.\ single-pass
F1=0.918$\pm$0.001). Two runs produced FP in the 500--700 range; one run
produced FP=1,524 with bimodal per-note prediction patterns (5--10 vs.\
150--220 predictions per note). This variability, regardless of its source,
underscores the architectural fragility of free-form Pass~2 rewriting.

The underlying failure mechanism is Pass~2's free-form text reformatting,
which accumulates alignment-based false positives. On the enhanced 10-note
gold, dual-pass scores F1=0.782 (all-configurations table, Supplement~1), reflecting this fundamental
limitation. Dual-pass recall on the enhanced gold (0.850) is essentially
identical to Baseline single-pass (0.850), despite a small recall
gain on the full corpus. The 10-note subset was selected for high
institutional PHI density, precisely the notes where Pass~2 triggers
mode-collapse (generating commentary or over-condensed output that fails
the validity check). The recall gain is concentrated in notes with moderate
institutional PHI density where Pass~2 operates normally.

The structured JSON output constraint (SP(Baseline) + Structured P2) eliminates this
failure mode by preventing the model from generating free-form redacted
text in Pass~2.

Multi-pass architectures have been reported to improve recall in settings where
notes are chunked before processing~\cite{redactor2025}. We attribute the
divergence from our findings to input length: their gains derive from
re-attending to chunks that do not fit a single context window, whereas our
notes are processed whole, so Pass~2 receives no information unavailable to
Pass~1.

\subsubsection*{E4.3: Self-Refinement Experiments}

The high variance of the Scrubber--Auditor~(Precision) pipeline (SD$_{\text{F1}}$=0.018;
Section~E4.1) suggests that some runs produce substantially suboptimal output. We
investigated whether lightweight self-refinement
(Self-Refine~\cite{madaan2023selfrefine},
Reflexion~\cite{shinn2023reflexion}) or ensemble strategies could stabilize
performance.

\paragraph{Self-Heal (Recall).} A Sonnet verifier examines the
Scrubber--Auditor~(Precision) output for residual PHI; if flagged, the full pipeline
retries with reflexion-style feedback. \textbf{Result:} F1=0.899,
R=0.958, P=0.846. The verifier flagged 2/10~notes, but the pipeline did
not outperform the base mean (F1=0.906) because the dominant variance
source is \emph{precision} instability (FP range 116--327), not recall
failures (FN range 75--90).

\paragraph{Self-Heal (Precision).} A Sonnet verifier reviews all
placeholders for over-redaction; if $\geq$4 are detected, only the Auditor
retries with a targeted restoration list. \textbf{Result:} F1=0.897,
R=0.958, P=0.844. The verifier triggered \emph{zero retries}: over-redactions
are distributed at 1--3 per note across ambiguous categories (institution
abbreviations, time-of-day values, pager numbers), never reaching the
threshold.

\paragraph{Same-Model Majority Vote (3$\times$ SA-Precision).} Three independent
runs of Scrubber--Auditor~(Precision); span-level majority vote ($\geq$2/3).
\textbf{Result:} F1=0.899, R=0.959, P=0.846. The three runs scored
F1$\in$[0.902, 0.906] with FP counts of 293, 286, and 296. Because
over-redactions are \emph{highly correlated} across runs (the same ambiguous
spans are consistently over-redacted), majority voting cannot filter them.

\paragraph{Cross-Model Majority Vote (Sonnet+DeepSeek+Kimi).} Three models
each run Precision independently; span-level majority vote ($\geq$2/3). These
models have diverse error profiles (Jaccard FN overlap 0.30--0.43).
\textbf{Result (3 trials):} Mean F1=0.902 (SD=0.006), R=0.986, P=0.832.
The ensemble achieves the highest recall of any configuration (0.986) but
precision \emph{decreases} relative to Sonnet alone (0.832 vs.\ the 5-trial
single-pass mean of 0.844) because weaker models contribute spurious
redactions that survive the 2/3~threshold.

\paragraph{Discussion.} No self-refinement or ensemble strategy improves
over Sonnet~(Precision) single-pass (mean F1=0.907, SD=0.002). Self-Heal verifiers
cannot detect the distributed, low-density over-redaction pattern that
drives FP variance. Same-model voting fails because over-redactions are
correlated across runs. Cross-model voting improves recall to near-ceiling
(0.986) but weaker models degrade precision, yielding no net F1 gain.
Self-refinement and ensemble costs ranged from 1.7$\times$ to 6$\times$ the
single-pass baseline (17--60 LLM calls vs.\ 10) without F1 improvement.

The fundamental insight is that the agentic pipeline's high-variance,
high-ceiling behavior (best run F1=0.942) cannot be reliably extracted
through simple post-hoc strategies. The rare optimal runs arise from
stochastic variation in the Auditor's restoration aggressiveness, a
property not amenable to consensus-based or verifier-based recovery.
For this task and evaluation set, well-calibrated prompt engineering on a
single strong model represents the practical performance ceiling.

\subsubsection*{E4.4: Cost and Latency}

Multi-agent pipelines require 2--6 LLM calls per note vs.\ 1 for single-pass:
2 for dual-pass and Scrubber--Auditor, 3 for cross-model voting, and 6 for
same-model voting (3~runs $\times$ the 2-call Scrubber--Auditor pipeline). The SP(Baseline) + Structured P2 pipeline reuses pre-computed Baseline
output, completing in $\sim$1.9~minutes for 100~notes (concurrent requests).
A full end-to-end dual-pass takes $\sim$90~minutes vs.\ $\sim$45~minutes for
single-pass. Self-refinement and ensemble strategies compound this overhead:
Self-Heal variants use 1.7$\times$ the base cost, while voting approaches
require 3--6$\times$ (30 LLM calls for cross-model vote, 60 for same-model
vote). At Sonnet~4.6 pricing (\$0.009/note for single-pass on the primary
benchmark), the Scrubber--Auditor pipeline adds Opus~4.8 costs
(\$0.044/note), making the agentic approach $\sim$6$\times$ more expensive
per note. Since none of the tested agentic configurations improved F1 over Precision single-pass,
the additional cost and latency are justified only when maximizing recall is
the sole priority or when using multi-agent workflows as a discovery tool
for gold standard refinement.



\subsection*{Section E5: Independent Validation Cost and Latency}

This section provides the per-system cost and latency breakdown for the
independent validation set, supporting the results summarized in Supplement~1,
Supplementary Note~4. The TiDE provisioning configurations themselves are
derived in Supplement~1, Supplementary Note~4.1.

\begin{table}[H]
\centering
\caption{Cost and latency comparison on the USDHUB
validation dataset (49~notes, sequential processing). LLM costs reflect
AWS Bedrock on-demand pricing; local systems have zero marginal API cost
(compute is pre-provisioned on the evaluation instance).}
\label{tab:usdhub_cost}
\begin{tabular}{lrrrr}
\toprule
\textbf{System} & \textbf{Latency} & \textbf{Cost (49 notes)} & \textbf{Cost/note} & \textbf{Platform} \\
\midrule
Claude Opus 4.8    & 6.8 min  & \$2.142 & \$0.0437 & Bedrock API \\
Claude Sonnet 4.6  & 4.6 min  & \$0.428 & \$0.0087 & Bedrock API \\
GPT-oss-120B       & 5.6 min  & \$0.400 & \$0.0082 & Bedrock API \\
DeepSeek V3.2      & 6.4 min  & \$0.166 & \$0.0034 & Bedrock API \\
Kimi K2.5          & 3.3 min  & \$0.166 & \$0.0034 & Bedrock API \\
GLM-5              & 11.5 min & \$0.166 & \$0.0034 & Bedrock API \\
MiniMax M2.5       & 17.3 min & \$0.166 & \$0.0034 & Bedrock API \\
GPT-oss-20B        & 61.7 min & \$0.043 & \$0.0009 & Bedrock API \\
\midrule
TiDE               & 27 s     & --- & --- & Local (Java) \\
OpenMed PII        & 45 s     & --- & --- & Local (434M) \\
spaCy NER + Regex  & 3.5 s    & --- & --- & Local (Python) \\
Regex-only         & $<$1 s   & --- & --- & Local (Python) \\
\bottomrule
\multicolumn{5}{l}{\small Latency measured end-to-end (sequential, single-threaded). LLM latency includes} \\
\multicolumn{5}{l}{\small network round-trip and API queuing; local systems include model loading.} \\
\multicolumn{5}{l}{\small Cost estimated from ${\sim}$45K input + ${\sim}$20K output tokens at published Bedrock rates.} \\
\end{tabular}
\end{table}

LLM API costs ranged from \$0.04 (GPT-oss-20B) to \$2.14 (Claude Opus~4.8)
for the full 49-note corpus, corresponding to \$0.0009--\$0.044 per note.
The highest-F1 system (Opus) is also the most expensive per note, while
three third-party models (DeepSeek, Kimi, GLM-5) cluster at
\$0.003/note, roughly 13$\times$ cheaper than Opus with recall within 3
percentage points (MiniMax clusters at similar cost but with 20-point
lower recall). Local systems (TiDE, spaCy, OpenMed, regex) incur zero
marginal API cost and run one to three orders of magnitude faster (3.5~s for
spaCy and 27~s for TiDE, versus 3.3--61.7~min for the LLMs).
GPT-oss-20B's anomalous latency (61.7~min) reflects persistent API errors
requiring retries, not model inference time.

\clearpage
\bibliography{references}

\clearpage
\section*{Extended Supplementary Figures}

\begin{suppfigure}[H]
    \centering
    \includegraphics[width=0.85\textwidth]{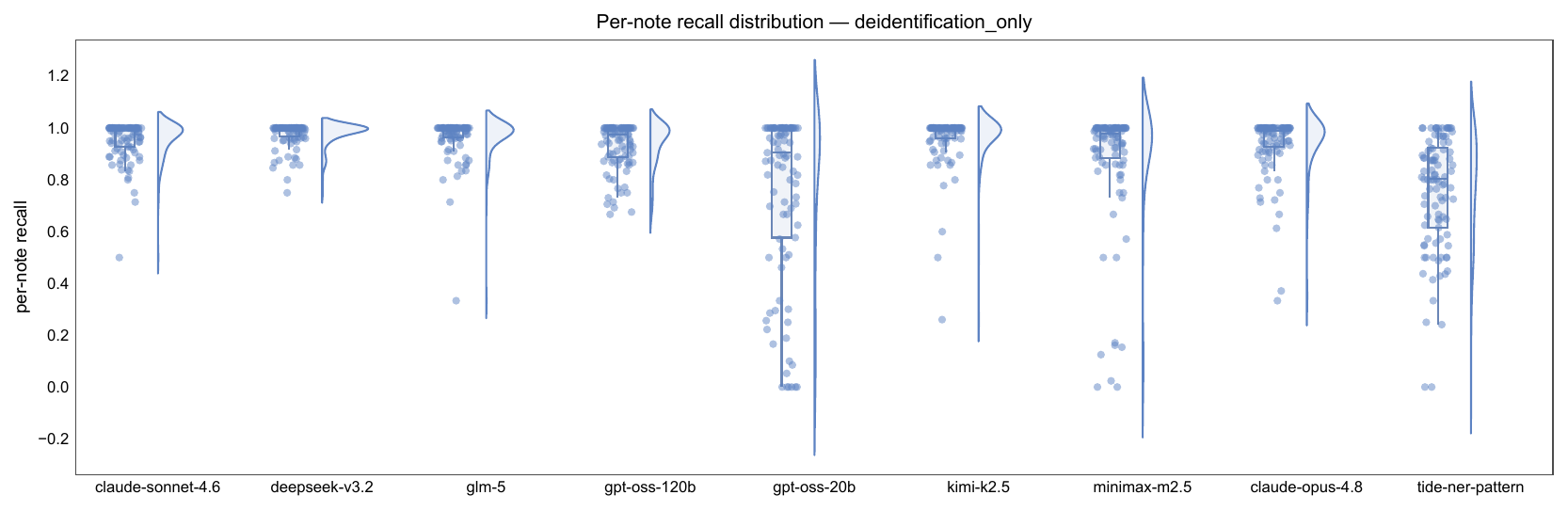}
    \caption{\textbf{Per-note recall distribution (Baseline).}
    Each point represents one clinical note. Most top models cluster near
    perfect recall (1.0), but all exhibit a tail of low-recall notes; these
    correspond to notes rich in institutionally situated PHI.}
    \label{suppfig:per_note_recall_v1}
\end{suppfigure}

\begin{suppfigure}[H]
    \centering
    \includegraphics[width=\textwidth]{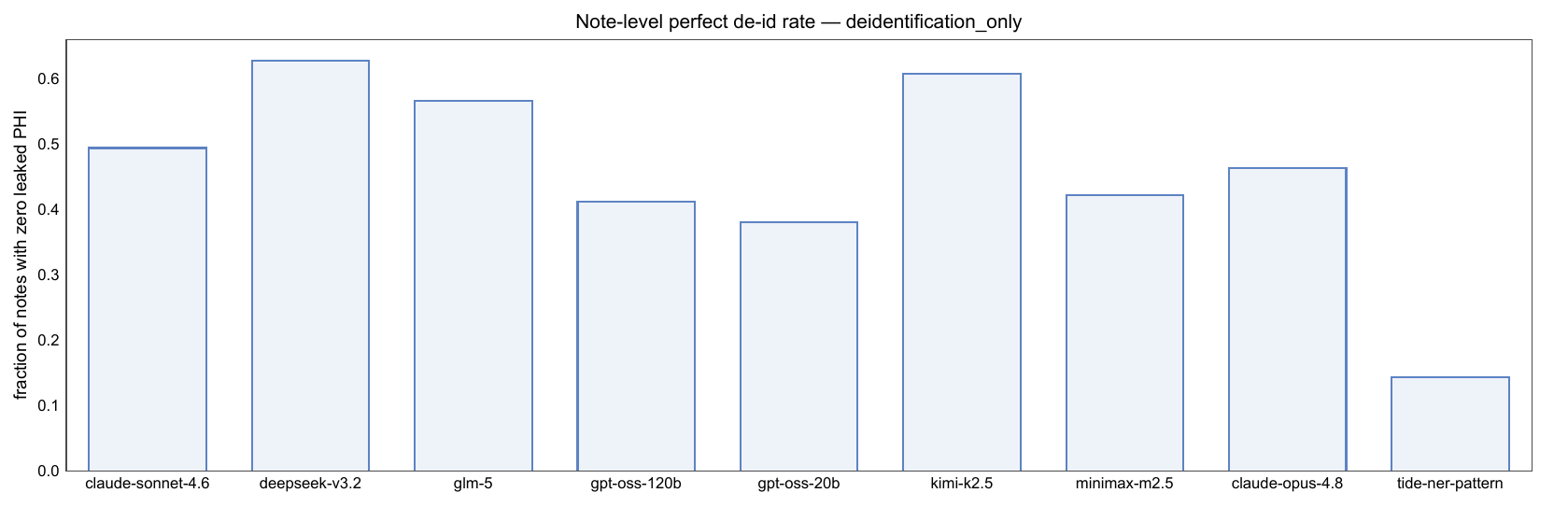}
    \caption{\textbf{Perfect de-identification rate (Baseline).}
    Fraction of notes with zero leaked PHI spans. DeepSeek and GLM-5 achieve
    perfect de-identification on $\sim$60\% of notes; no model exceeds 65\%,
    reflecting the long tail of institutionally situated identifiers.}
    \label{suppfig:perfect_deid_rate_v1}
\end{suppfigure}

\begin{suppfigure}[H]
    \centering
    \includegraphics[width=\textwidth]{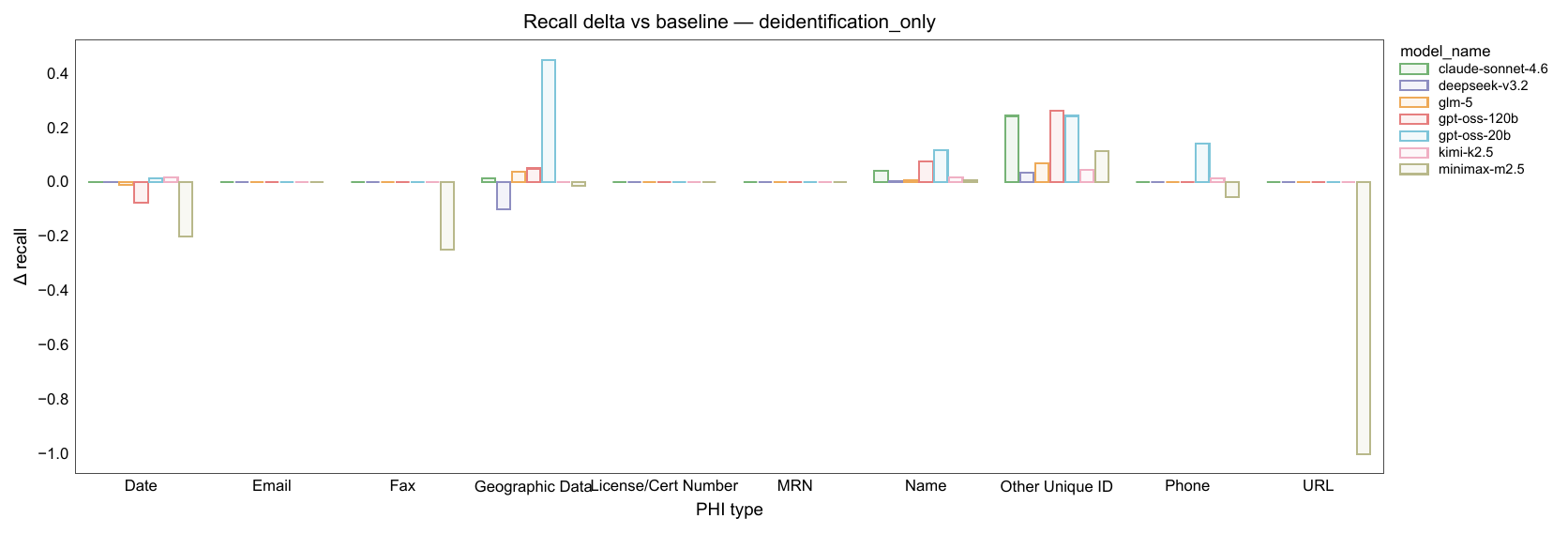}
    \caption{\textbf{Per-category recall change under targeted prompting
    (Targeted~$-$~Baseline).} The targeted prompt yields concentrated gains in Other
    Unique~ID and Name (institutionally situated categories) for responsive
    models, while MiniMax~M2.5 shows broad regressions. Geographic Data shows
    mixed effects.}
    \label{suppfig:recall_delta_v2}
\end{suppfigure}

\begin{suppfigure}[H]
    \centering
    \includegraphics[width=\textwidth]{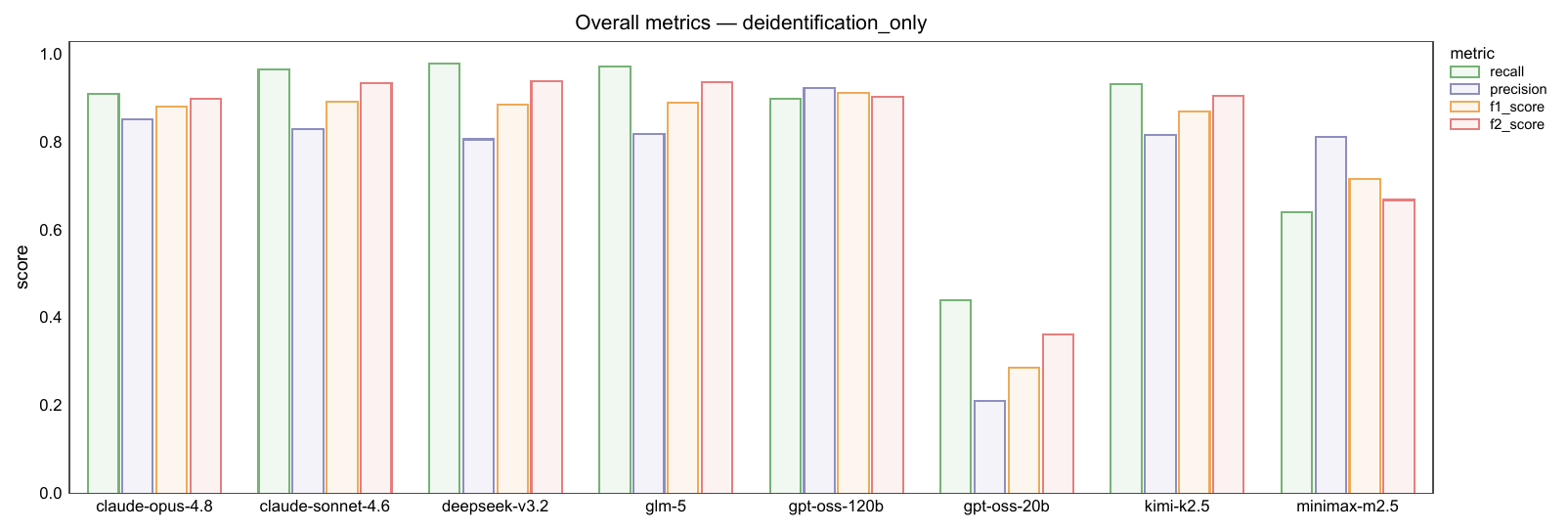}
    \caption{\textbf{Overall performance under precision-focused prompt (Precision).}
    The "do-not-over-redact" instructions improve precision across most models
    relative to Baseline/Targeted, with GPT-oss-120B showing the most dramatic improvement.
    GPT-oss-20B suffers output failures (reasoning exhausts token budget).}
    \label{suppfig:overall_bars_v3}
\end{suppfigure}

\begin{suppfigure}[H]
    \centering
    \includegraphics[width=\textwidth]{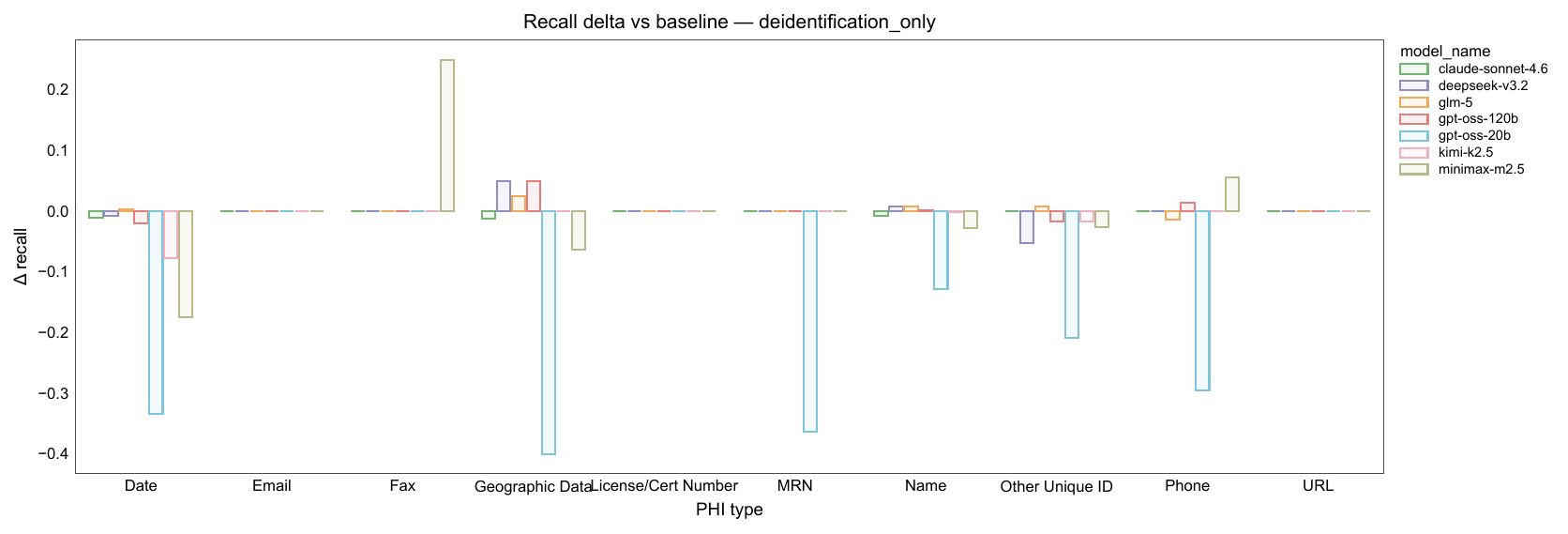}
    \caption{\textbf{Per-category recall change (Precision~$-$~Baseline).}
    Unlike Targeted (Supplementary Figure~\ref{suppfig:recall_delta_v2}), the Precision
    prompt shows smaller recall deltas for strong models, confirming that
    precision can be improved without substantial recall degradation. GPT-oss-20B
    and MiniMax~M2.5 are outliers with large regressions.}
    \label{suppfig:recall_delta_v3}
\end{suppfigure}

\begin{suppfigure}[H]
    \centering
    \includegraphics[width=\textwidth]{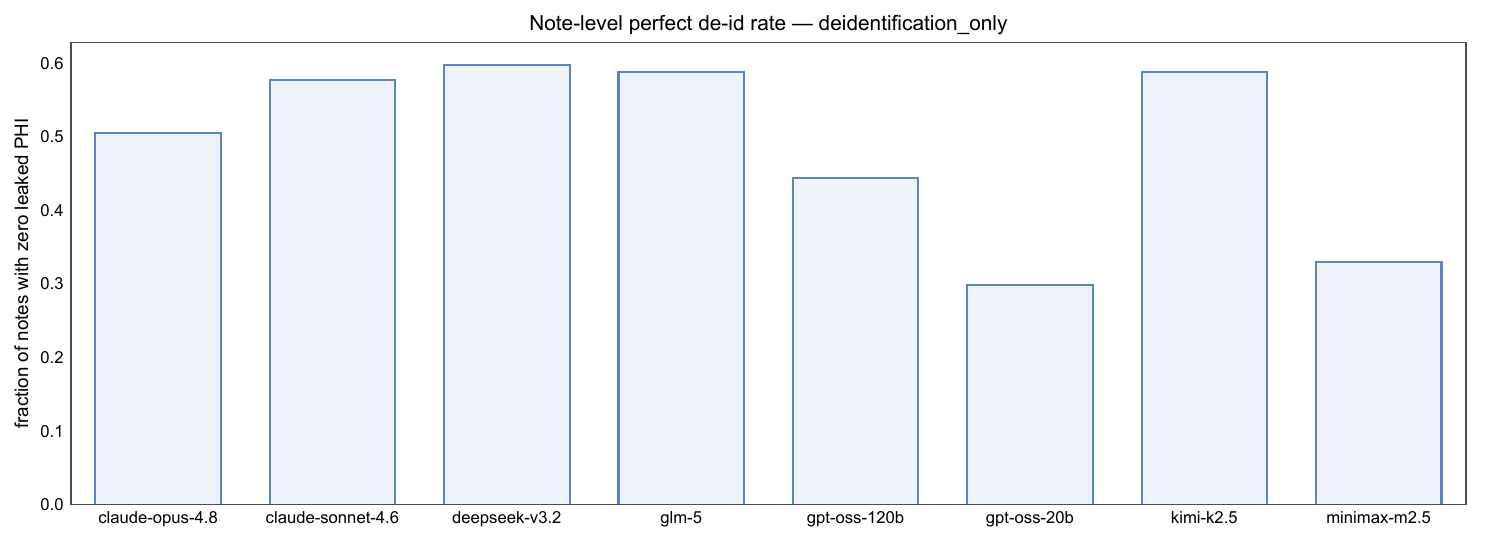}
    \caption{\textbf{Perfect de-identification rate (Precision).}
    Under the precision-focused prompt, DeepSeek V3.2 and Kimi K2.5 maintain
    the highest perfect de-id rates ($\sim$60\%), while GPT-oss-20B and
    MiniMax~M2.5 show substantial degradation. Compare with
    Supplementary Figure~\ref{suppfig:perfect_deid_rate_v1} for baseline.}
    \label{suppfig:perfect_deid_rate_v3}
\end{suppfigure}

\begin{suppfigure}[H]
    \centering
    \includegraphics[width=0.85\textwidth]{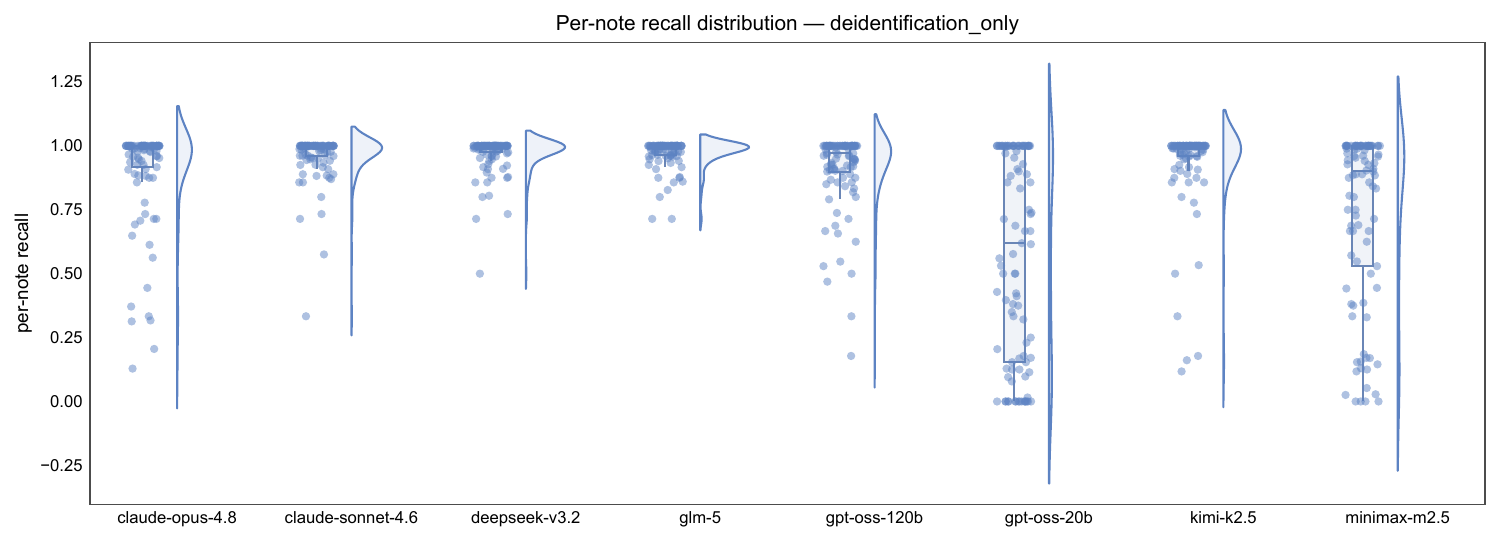}
    \caption{\textbf{Per-note recall distribution (Precision).}
    Compared to Baseline (Supplementary Figure~\ref{suppfig:per_note_recall_v1}),
    the distribution remains concentrated near 1.0 for strong models but shows
    increased variance for GPT-oss-20B and MiniMax~M2.5 under the
    precision-focused prompt.}
    \label{suppfig:per_note_recall_v3}
\end{suppfigure}

\begin{suppfigure}[H]
    \centering
    \includegraphics[width=0.7\textwidth]{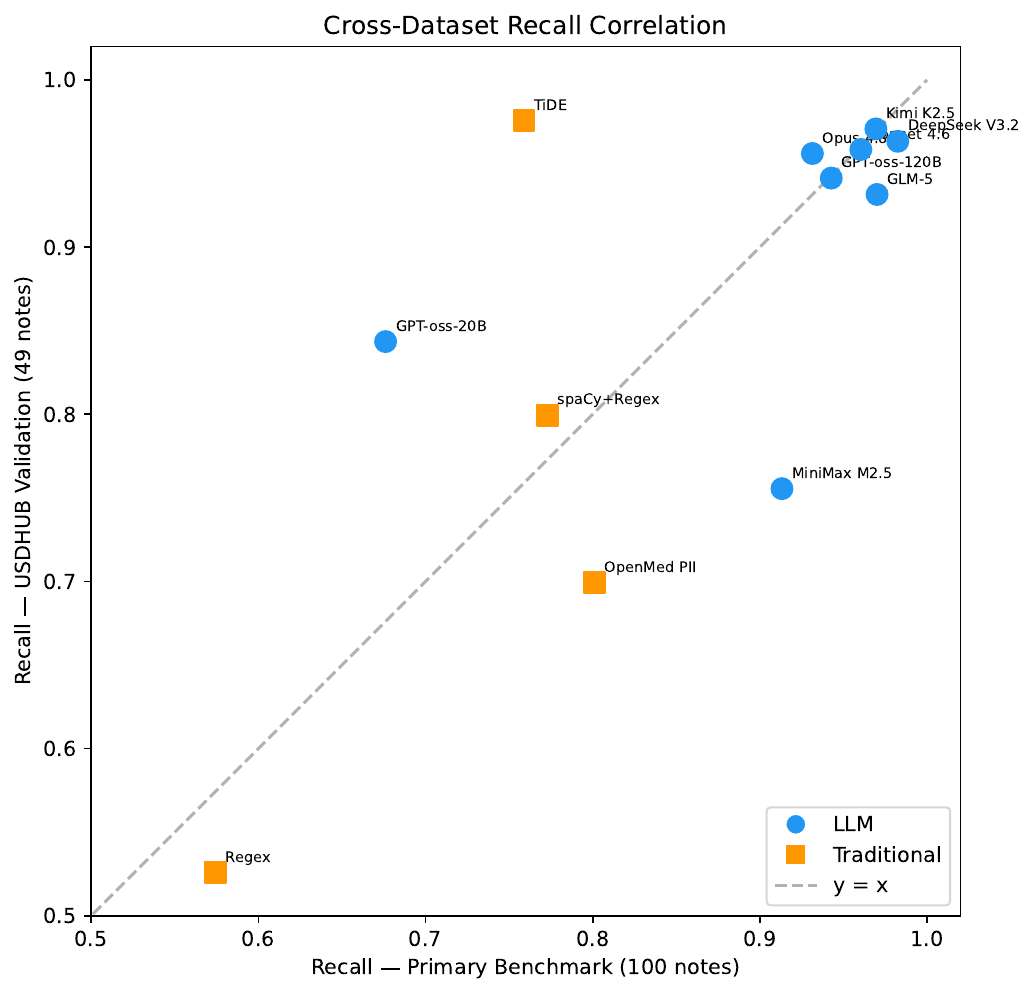}
    \caption{\textbf{Cross-dataset recall correlation.}
    Each point is one system; x-axis is recall on the primary benchmark,
    y-axis is recall on USDHUB. Points near the diagonal indicate consistent
    performance across corpora. TiDE shows the largest positive deviation
    (higher recall on USDHUB than primary). On USDHUB, TiDE was additionally
    run with known-PHI matching enabled (a supplied per-note dictionary of
    identifiers; see Supplement~1, Supplementary Note~4.1), a near-oracle configuration not available
    to the LLMs, which accounts for much of this deviation; USDHUB also contains
    standard demographic PHI without the institutionally situated identifiers
    that challenge pattern-based systems.}
    \label{suppfig:recall_correlation}
\end{suppfigure}

\begin{suppfigure}[H]
    \centering
    \includegraphics[width=0.85\textwidth]{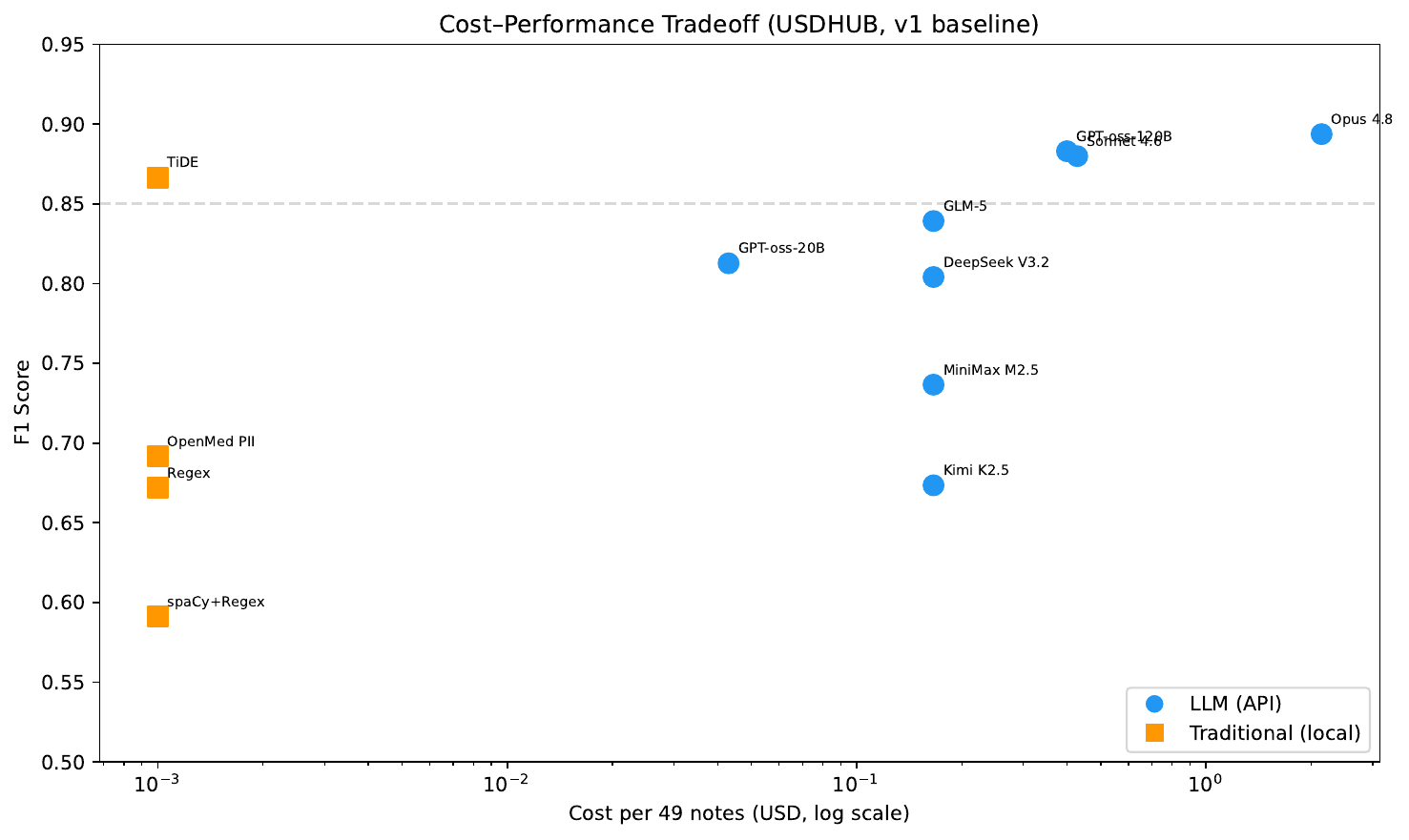}
    \caption{\textbf{Cost--performance tradeoff (USDHUB, Baseline).}
    F1 score vs.\ total API cost for the 49~notes (log scale). Traditional
    baselines (orange squares) incur zero marginal API cost. Among LLMs (blue
    circles), Opus~4.8 achieves the highest F1 at the highest cost
    (\$2.14 for the corpus, \$0.044/note), while DeepSeek, Kimi, GLM-5 and
    MiniMax all cost \$0.17 for the corpus (\$0.003/note) at F1~0.67--0.84. The
    cost--F1 Pareto frontier is GPT-oss-20B (\$0.04), GLM-5 (\$0.17),
    GPT-oss-120B (\$0.40) and Opus~4.8 (\$2.14); DeepSeek, Kimi and MiniMax are
    dominated. GPT-oss-20B is cheapest only because reasoning-token exhaustion
    truncated some outputs, so it is not a usable operating point despite lying
    on the frontier.}
    \label{suppfig:cost_f1}
\end{suppfigure}

\begin{suppfigure}[H]
    \centering
    \includegraphics[width=0.85\textwidth]{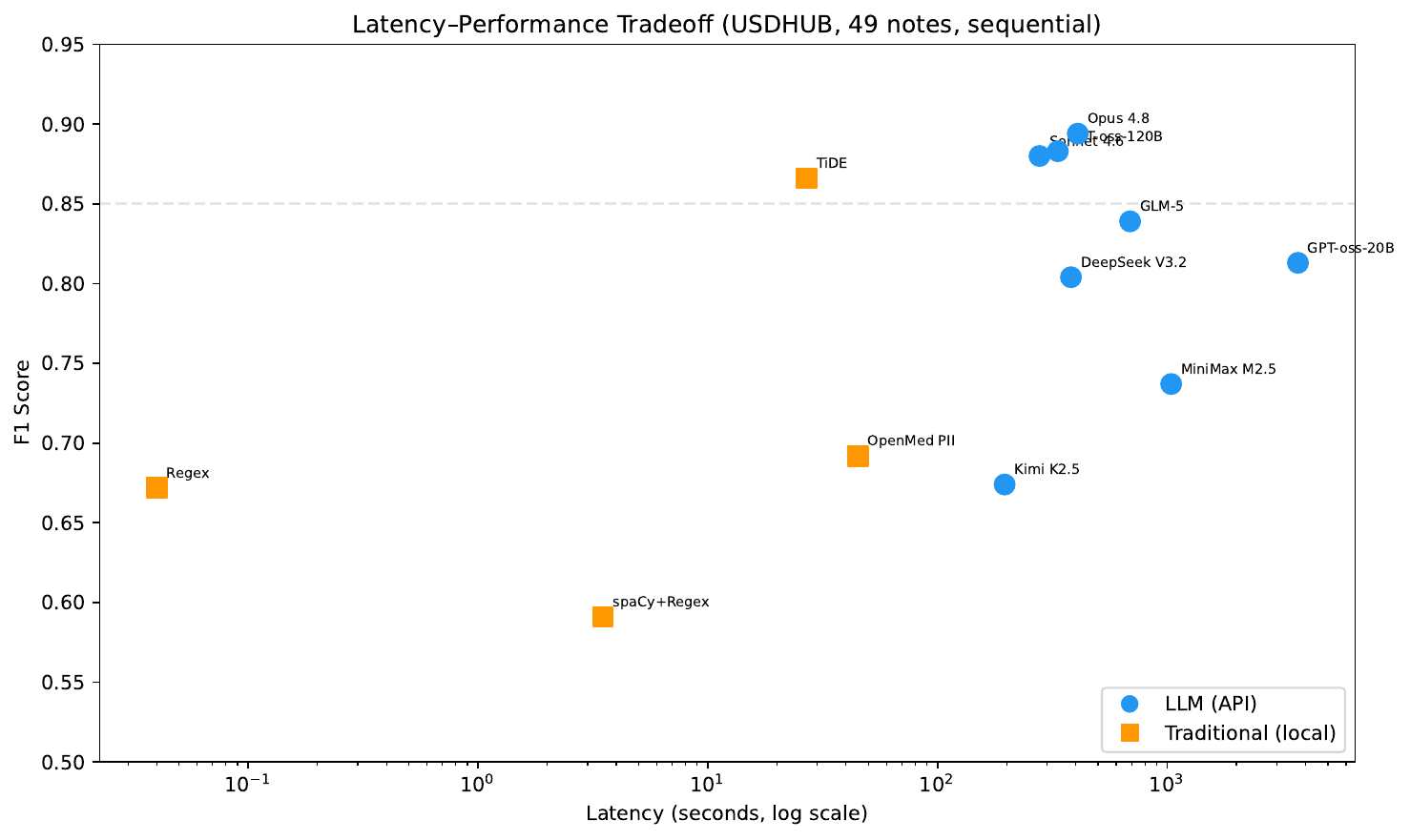}
    \caption{\textbf{Latency--performance tradeoff (USDHUB, 49~notes, sequential).}
    End-to-end processing time vs.\ F1. Local systems (TiDE, spaCy, OpenMed,
    regex) complete in under 1~minute with no API dependency. LLMs range from
    3--62~minutes depending on model and API stability. GPT-oss-20B's 62-minute
    runtime reflects retry overhead from persistent API errors, not inference
    speed.}
    \label{suppfig:latency_f1}
\end{suppfigure}

\begin{suppfigure}[H]
    \centering
    \includegraphics[width=\textwidth]{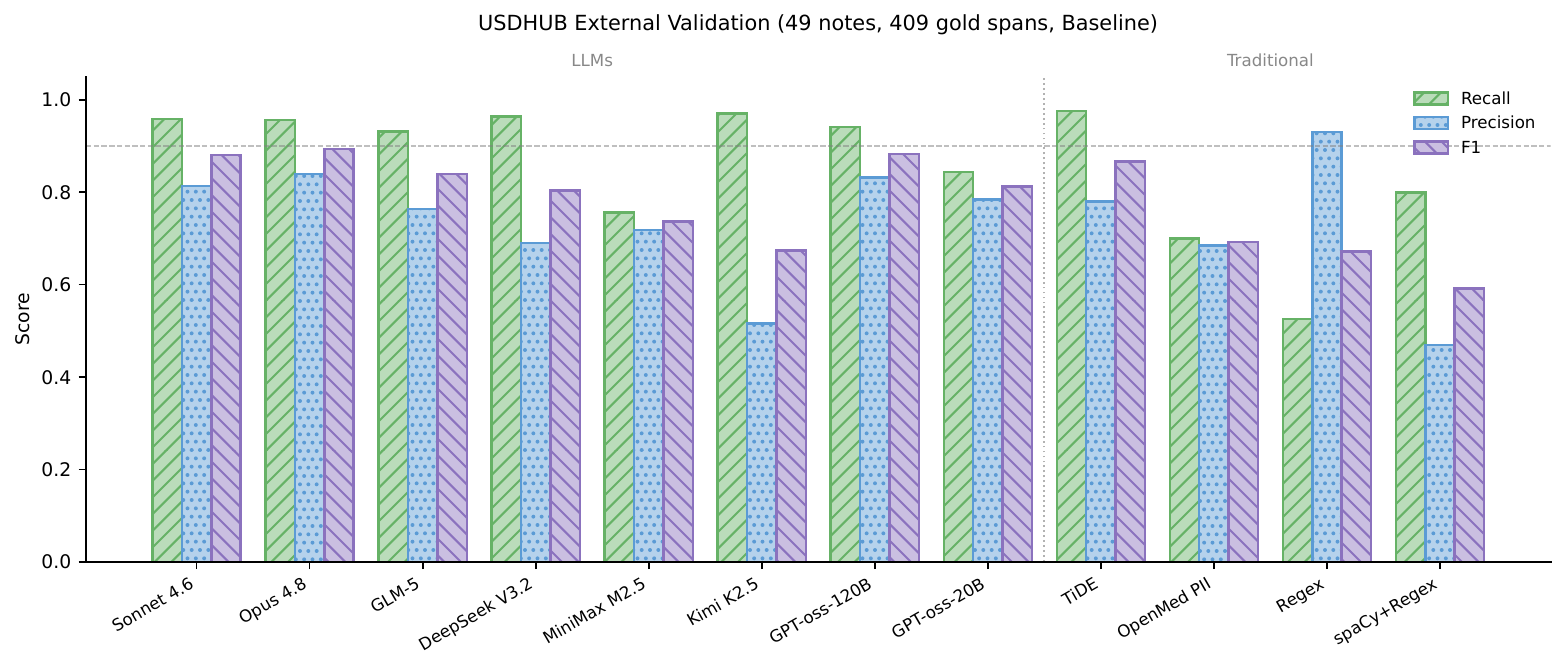}
    \caption{\textbf{Overall performance on USDHUB validation
    (49~notes, 409~gold spans, Baseline).} LLMs are evaluated under identical
    conditions as the primary benchmark. TiDE achieves the highest recall
    (0.976 fully unprovisioned; 0.971 maximally provisioned with a known-PHI
    dictionary and per-note header, a difference of $<$0.6~recall points on this
    canonical-PHI corpus; Supplement~1, Supplementary Note~4.1) while Opus~4.8 leads on F1
    (0.894). The LLM--traditional gap is clearly visible, with traditional
    methods clustered below F1=0.70 except TiDE.}
    \label{suppfig:usdhub_overall}
\end{suppfigure}

\begin{suppfigure}[H]
    \centering
    \includegraphics[width=0.9\textwidth]{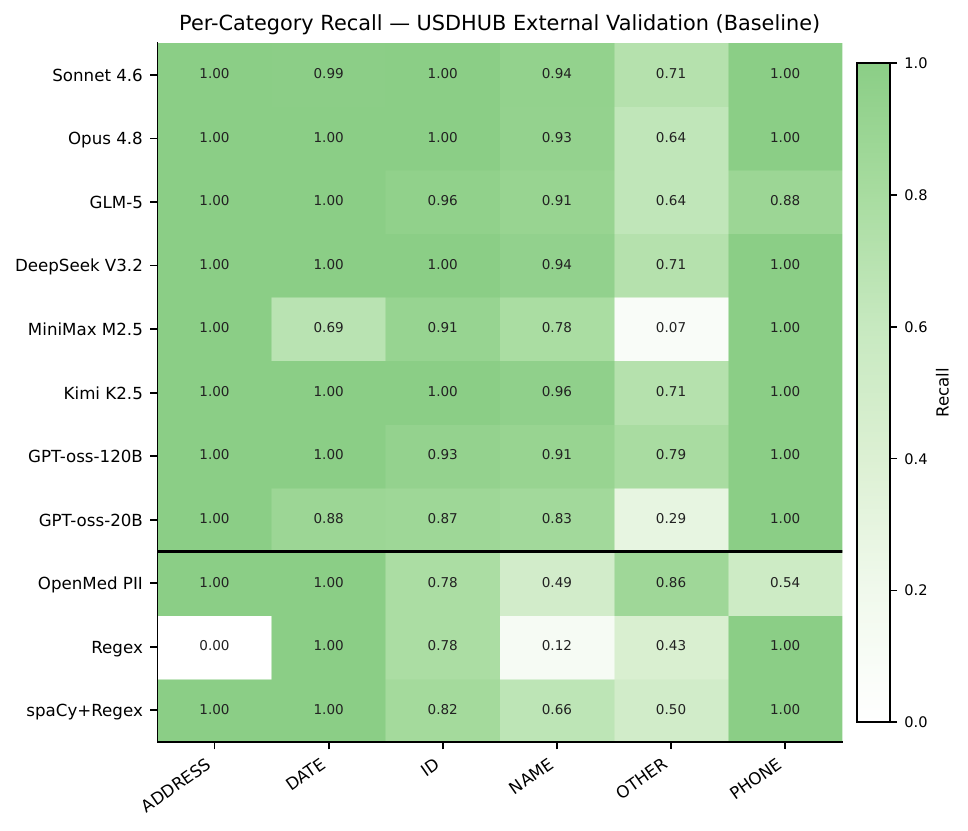}
    \caption{\textbf{Per-category recall heatmap, USDHUB validation.}
    PHI categories in USDHUB (Name, Date, ID, Phone, Address, Other) show
    similar patterns to the primary benchmark: LLMs achieve near-perfect recall
    on most categories while traditional baselines show systematic gaps,
    particularly on Names.}
    \label{suppfig:usdhub_heatmap}
\end{suppfigure}

\begin{suppfigure}[H]
    \centering
    \includegraphics[width=\textwidth]{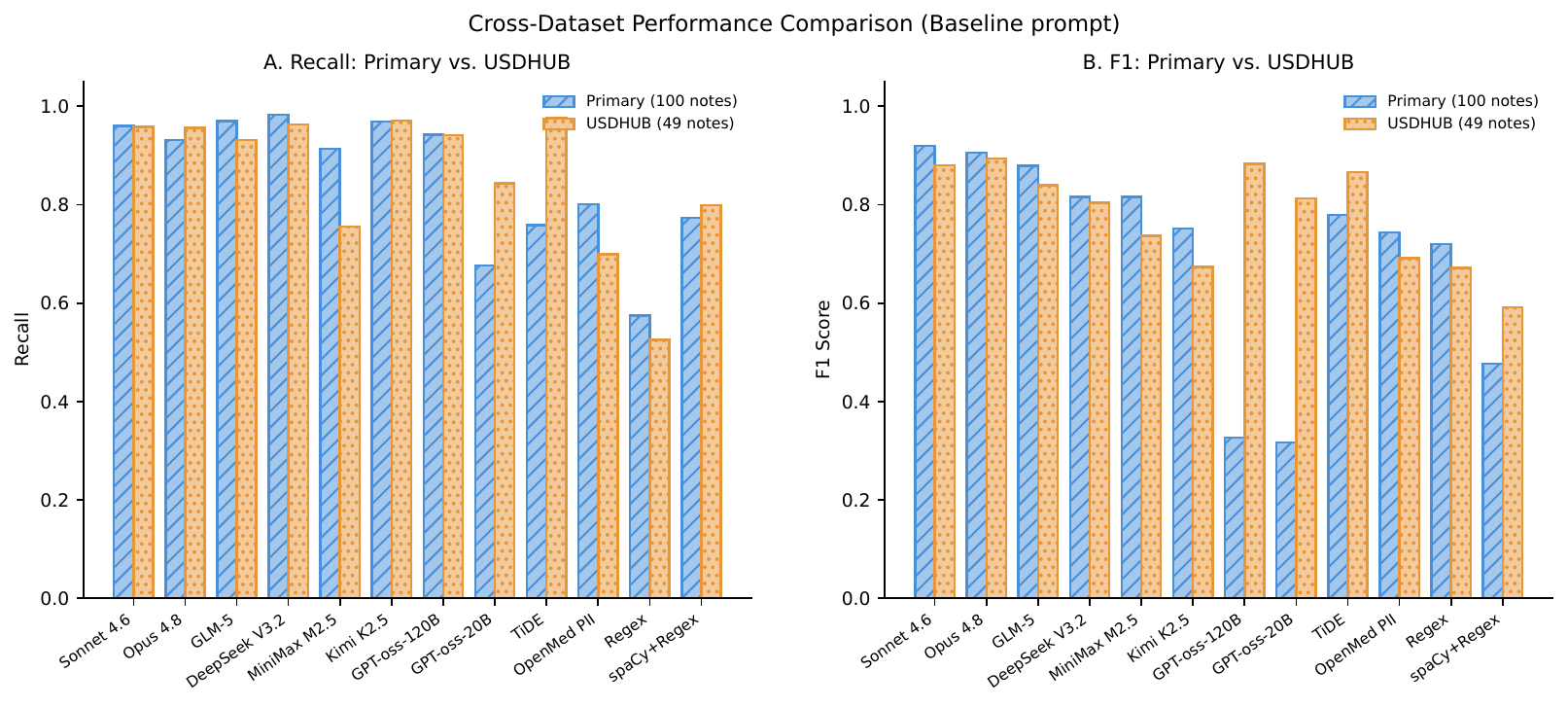}
    \caption{\textbf{Cross-dataset performance comparison.}
    Side-by-side recall (A) and F1 (B) for all systems on the primary benchmark
    (100~notes, blue) vs.\ USDHUB validation (49~notes, orange). Both datasets
    use the Baseline prompt. Performance rankings are largely preserved across
    datasets, supporting generalizability of the findings.}
    \label{suppfig:cross_dataset}
\end{suppfigure}